%% file: main.tex
\documentclass{article} 
\usepackage{iclr2025_conference,times}

\usepackage{hyperref}
\usepackage{url}
\usepackage{wrapfig}
\usepackage{color}
\usepackage{tabularx}
\usepackage{tcolorbox}
\usepackage{xcolor}
\usepackage{colortbl}
\usepackage{tabu}
\usepackage{booktabs}
\usepackage{multirow}

\usepackage{graphicx} 
\usepackage{float}
\usepackage{subfigure} 
\usepackage{amssymb}

\usepackage{tablefootnote}
\usepackage{xspace}

\usepackage{float}
\usepackage{amsmath}
\usepackage{cleveref}
\usepackage{bm}
\usepackage{algorithmicx}
\usepackage{algorithm}
\usepackage{algpseudocode}

\usepackage{arydshln}

\usepackage{amssymb}
\usepackage{pifont}
\newcommand{\cmark}{\textcolor{green!35!black}{\ding{51}}}
\newcommand{\xmark}{\ding{55}}%
\usepackage{enumitem}
\usepackage{bm}

\usepackage{cuted}

\usepackage{gradient-text}
\usepackage{fontawesome5}

\usepackage{caption}
\usepackage{array}
\usepackage[subfigure]{tocloft}
\usepackage{makecell}

\usepackage{fancyhdr}

\usepackage{listings}
\addtocontents{toc}{\protect\setcounter{tocdepth}{-1}}

\newcommand{\projectlogo}{\raisebox{-1.8pt}{\includegraphics[height=1.3em]{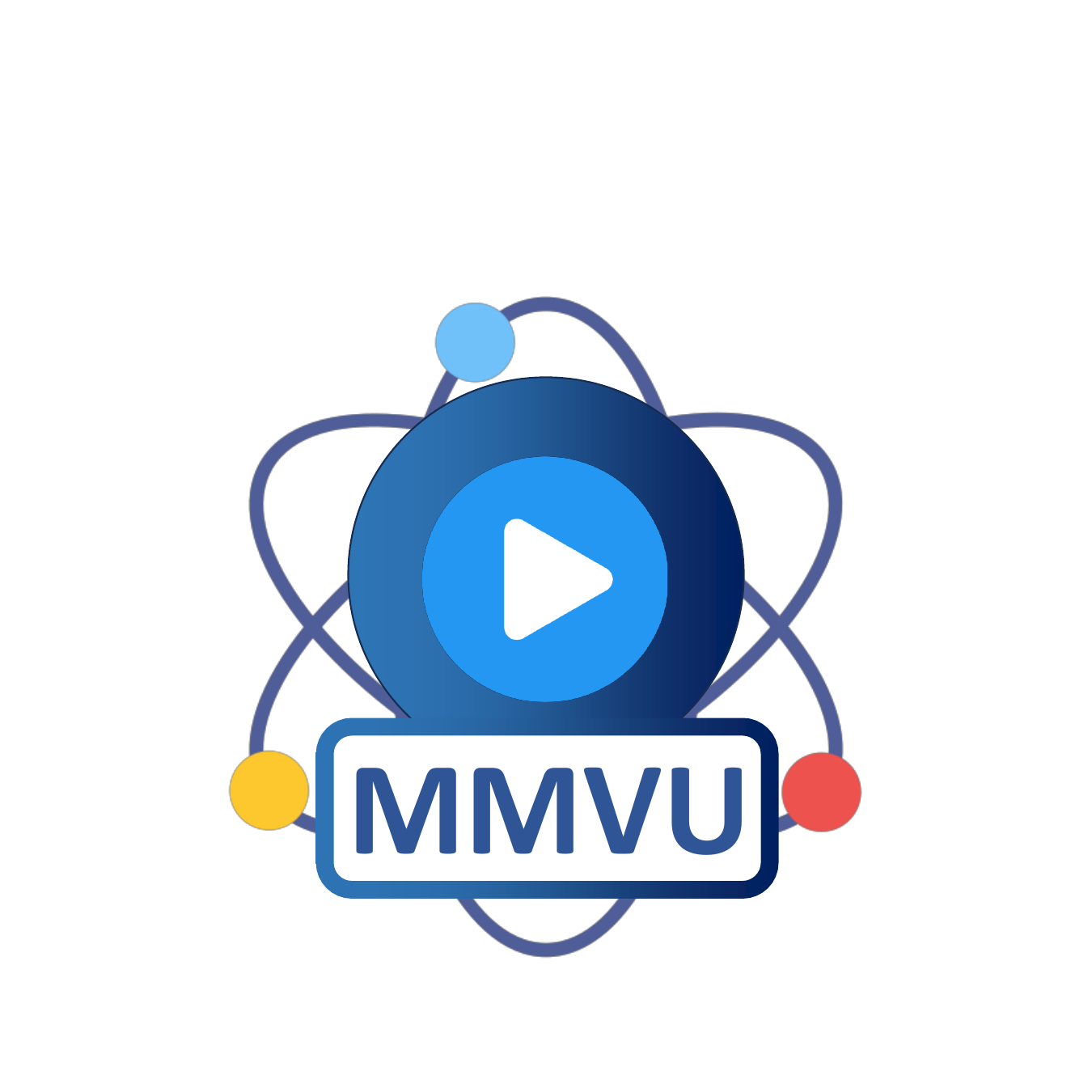}}\xspace}
\newcommand{\huggingface}{\raisebox{-1.5pt}{\includegraphics[height=1.05em]{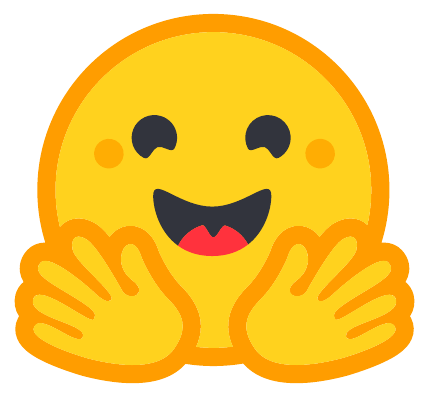}}\xspace}
\newcommand{\github}{\raisebox{-1.5pt}{\includegraphics[height=1.05em]{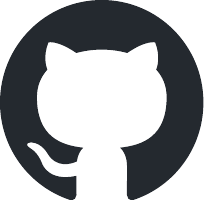}}\xspace}

\newcommand{\ours}{\gradientRGB{MMVU}{53,93,203}{10,10,80}\xspace}
\renewcommand{\cite}{\citep}

\newcommand{\titleLogo}{%
    $\raisebox{-2.8mm}
    {\includegraphics[width=0.08\textwidth]{icons/MMVU.pdf}}$ 
}

\newcommand{\multidis}{multi-discipline\xspace}

\newcommand{\eg}{\hbox{\emph{e.g.,}}\xspace}
\newcommand{\ie}{\hbox{\emph{i.e.,}}\xspace}
\newcommand{\gpt}{66.7\%\xspace}
\newcommand{\mmworldrate}{39.5\%}

\newcommand{\da}{Direct Answer\xspace}
\newcommand{\nsubjects}{27\xspace}
\newcommand{\nexample}{3,000\xspace}
\newcommand{\nerror}{six\xspace}
\newcommand{\nval}{1,000\xspace}
\newcommand{\ntest}{2,000\xspace}
\newcommand{\nmodel}{32\xspace}
\newcommand{\norg}{17\xspace}
\newcommand{\nvideo}{1,529\xspace}

\usepackage{multicol}

\title{\titleLogo\textbf{\ours}: Measuring Expert-Level Multi-Discipline Video Understanding}

\iclrfinalcopy 
\begin{document}

\author{Yilun Zhao\thanks{Core Contributors. All authors’ contributions are detailed in the Contribution section.} 
~~ Lujing Xie$^*$
~Haowei Zhang$^*$
~Guo Gan 
~ Yitao Long
~ Zhiyuan Hu
~ Tongyan Hu 
\\
{\bf 
Weiyuan Chen
~ Chuhan Li
~ Junyang Song 
~ Zhijian Xu 
~ Chengye Wang 
~ Weifeng Pan
}\\
{\bf 
Ziyao Shangguan 
~ Xiangru Tang 
~ Zhenwen Liang 
~ Yixin Liu
~ Chen Zhao 
~ Arman Cohan
}\\
[6pt]
\makebox[\textwidth]{\centering {Yale NLP MMVU Team}}
\vspace{-10pt}
}

\maketitle
\begin{abstract}
\input{main/0-abstract}
\end{abstract}

\input{main/1-introduction}
\input{main/2-related_work}
\input{main/3-dataset}
\input{main/4-experiment}

\input{main/5-conclusion}

\bibliography{custom, llm, textbook}
\bibliographystyle{iclr2025_conference}

\clearpage
\appendix

\addtocontents{toc}{\protect\setcounter{tocdepth}{3}}

\hypersetup{linkcolor=black}

\renewcommand{\contentsname}{\large Appendix Contents}
\let\oldtableofcontents\tableofcontents

\renewcommand{\tableofcontents}{%
    \begingroup
    \setlength{\cftbeforesubsecskip}{2pt} 
    \hypertarget{toc}{}
    \oldtableofcontents
    \endgroup
}

\tableofcontents

\clearpage

\pagestyle{fancy}
\renewcommand{\headrulewidth}{0pt}
\fancyhead{}
\fancyfoot[R]{\hyperlink{toc}{Back to Appendix Table of Contents}}

\clearpage
\input{appendix/A-dataset_construction}

\input{appendix/B-experiment_setup}

\input{appendix/E-case_study}

\end{document}

%% file: main/0-abstract.tex
We introduce \ours, a comprehensive expert-level, \multidis benchmark for evaluating foundation models in video understanding. 
\ours includes \nexample expert-annotated questions spanning \nsubjects subjects across four core disciplines: Science,  Healthcare, Humanities \& Social Sciences, and Engineering.
Compared to prior benchmarks, 
\ours features three key advancements.
First, it challenges models to apply domain-specific knowledge and perform expert-level reasoning to analyze specialized-domain videos, moving beyond the basic visual perception typically assessed in current video benchmarks.
Second, each example is annotated by human experts from scratch. We implement strict data quality controls to ensure the high quality of the dataset.
Finally, each example is enriched with expert-annotated reasoning rationals and relevant domain knowledge, facilitating in-depth analysis.
We conduct an extensive evaluation of \nmodel frontier multimodal foundation models on \ours. 
The latest System-2-capable models, o1 and Gemini 2.0 Flash Thinking, achieve the highest performance among the tested models. However, they still fall short of matching human expertise. Through in-depth error analyses and case studies, we offer actionable insights for future advancements in expert-level, knowledge-intensive video understanding for specialized domains.

\vspace{-10pt}
\renewcommand{\arraystretch}{1.2}
\begin{center}
\begin{tabular}{cll}
\projectlogo & \textbf{Project Page:} & \href{https://mmvu-benchmark.github.io}{\path{mmvu-benchmark.github.io}}\\
\huggingface & \textbf{\ours Data:} &  \href{https://huggingface.co/datasets/yale-nlp/MMVU} {\path{huggingface.co/datasets/yale-nlp/MMVU}}\\
\github & \textbf{\ours Code:} & \href{https://github.com/yale-nlp/MMVU}{\path{github.com/yale-nlp/MMVU}}\\
\end{tabular}
\end{center} 
\vspace{-12pt}

%% file: main/1-introduction.tex
\input{figures_tex/main_example}
\section{Introduction}
Foundation models have demonstrated remarkable capabilities in reasoning across various domains, yet their ability to handle expert-level knowledge remains a critical area of evaluation~\cite{hendrycks2021mmlu, yue2024mmmu}. 
In recent years, researchers have developed numerous benchmarks to assess these models' proficiency in specialized domains, primarily focusing on text-based reasoning~\cite{hendrycks2021mmlu, wang2024mmlupro, feng2024sciknoweval, scieval} and image-based contexts~\cite{lu2024mathvista, yue2024mmmu, yue2024mmmupro, zhang2024cmmmu, li2024mmsci}.
However, as capabilities of foundation models expand across multiple modalities, there is a significant gap in evaluating expert-level reasoning over specialized-domain \emph{videos}. This gap is particularly concerning as video is one of the most information-rich and naturalistic modalities, and is widely used to convey complex, dynamic information in specialized fields like healthcare, engineering, and scientific research~\cite{he2024mmworld}. 
Unlike static text or images, expert-level videos often capture temporal dynamics, procedural knowledge, and complex interactions that are essential in many specialized domains.
For example, in science, expert-level and knowledge-intensive reasoning might involve analyzing a chemical reaction video (\Cref{fig:overview}). 
A model must identify key reaction stages based on subtle visual cues like color changes or the formation of precipitates, which requires integrating chemical knowledge in addition to recognizing visual patterns. 

To bridge this gap, we introduce \ours, a comprehensive benchmark measuring \textbf{\underline{M}}ultimodal foundation models in expert-level, \textbf{\underline{M}}ulti-discipline \textbf{\underline{V}}ideo \textbf{\underline{U}}nderstanding and reasoning. 
\ours consists of \nexample expert-annotated QA examples over \nvideo specialized-domain videos, 
spanning \nsubjects subjects across four key disciplines: Science,  Healthcare, Humanities \& Social Sciences, and Engineering.
To ensure both the breadth of domain knowledge and the depth of reasoning required for \ours, we implement a textbook-guided data annotation process.
Expert annotators first locate key concepts from textbooks in their fields, then source relevant videos and create corresponding questions that require domain knowledge and expert-level reasoning to comprehend the videos. 
Each example also includes expert-annotated reasoning rationale and relevant domain knowledge, facilitating fine-grained evaluation of model performance.
Thorough data quality controls are implemented to ensure high quality of \ours. 

We conduct an extensive evaluation on \ours, covering \nmodel frontier multimodal foundation models from \norg organizations. 
Notably, the latest o1 model demonstrates the highest performance among all tested models, approaching the expertise of human experts. 
Despite this progress, other models still fall noticeably short of human-level capabilities. For instance, GPT-4o achieves a score of \gpt, which is substantially lower than the benchmark set by human experts (\ie, 86.8\%) in the open-book setting.
Our analysis highlights the effectiveness of CoT reasoning, which generally enhances model performance compared to directly generating final answers without intermediate reasoning steps.
To deepen understanding of the current models' limitations, we perform an in-depth error analysis of frontier models, including numerous case studies reviewed by human experts. These insights provide valuable guidance for future advancements in the field.

%% file: figures_tex/main_example.tex
\begin{figure}[h]
    \centering
    \includegraphics[width=\textwidth]{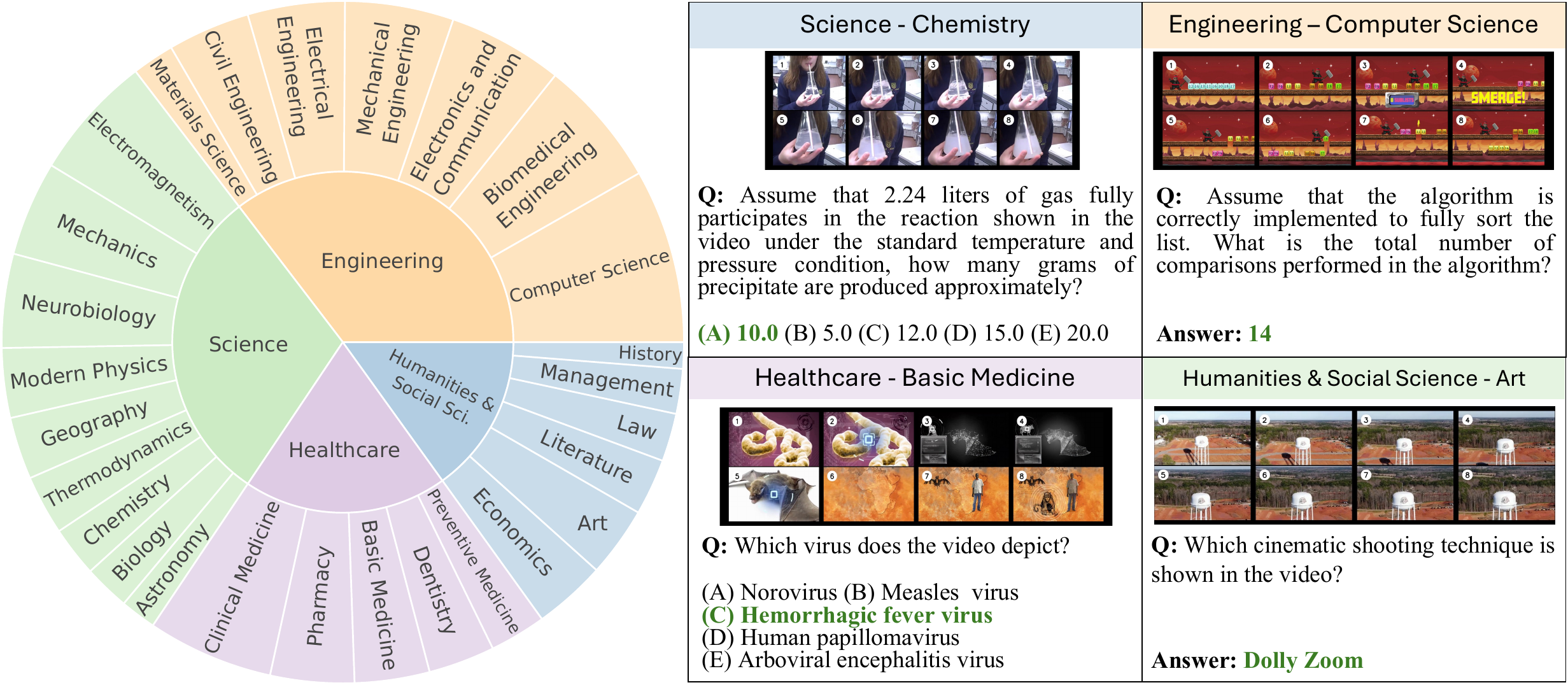}
    \captionof{figure}{
    Overview of the \ours benchmark. \ours includes \nexample expert-annotated examples, covering 27 subjects across four core disciplines. It is specifically designed to assess multimodal foundation models in expert-level, knowledge-intensive video understanding and reasoning tasks.
    }
    \label{fig:overview}
\end{figure}

%% file: main/2-related_work.tex
\section{Related Work}
\paragraph{Video Understanding Benchmark.}
Existing video understanding benchmarks primarily focus on \emph{general-purpose} video comprehension tasks, such as action recognition~\cite{7298698, sigurdsson2016hollywood,3365212,Deng_2023_ICCV}, captioning and description~\cite{xu2016msr, krishna2017dense, li2024mvbenchcomprehensivemultimodalvideo,takahashietal2024abstractive,BENCHMARKS2021_5ef05993}, grounding~\cite{lei-etal-2018-tvqa, wang2022geb+,Chen_2023_CVPR,kesen2023vilmazeroshotbenchmarklinguistic}, temporal reasoning~\cite{jang2017tgif, liu-etal-2024-tempcompass, shangguan2024tomato, cores2024tvbench, cai2024temporalbench, kesen2024vilma, li2024vitatecs}, and long video understanding~\cite{zhang2023movqa, wang2024lvben, neptune24, ataallah2024infinibench, fang2024mmbenchvideo}.
The rise of video-based foundation models~\cite{vidllmsurvey, zhang-etal-2023-video, fei2024videoccam, Huang_2024_CVPR} has driven the development of new benchmarks that include diverse video comprehension tasks for more comprehensive evaluation~\cite{xiao2021next, ning2023videobench, li2024mvbench, fu2024videomme, li2024videoeval, khattak2024goodvideo, yang2024thinking}. 
However, these benchmarks remain predominantly focused on natural scenes and general-purpose tasks.
A significant gap persists in benchmarks targeting \emph{expert-level} and \emph{knowledge-intensive reasoning} over specialized-domain videos, where both visual perception and domain-specific expertise are required—especially in critical fields like healthcare, engineering, and science~\cite{he2024mmworld}. 

\input{tables/data_comparison_table}
\paragraph{Multi-discipline Evaluation Benchmark.}
The rapid development of foundation models has significantly enhanced expert-level reasoning across various disciplines~\cite{Touvron2023Llama2O, Jiang2023Mistral7, yang2024qwen2, geminiteam2024gemini, openai2024gpt4o}.
Early benchmarks focused on domain-specific tasks for textual domains, establishing a foundation for assessing the models' strengths and limitations in expert reasoning~\cite{welbl-etal-2017-crowdsourcing, clark2018think, hendrycks2021mmlu, suzgun-etal-2023-challenging, zhong-etal-2024-agieval, chen-etal-2023-theoremqa, wang2024mmlupro, zhao-etal-2024-knowledgefmath}. 
More recently, benchmarks have evolved to include multimodal tasks~\cite{yue2024mmmu, lu2024mathvista, zhang2024cmmmu, yue2024mmmupro, li2024mmsci, wang2024charxiv}, emphasizing visual perception and advanced reasoning with domain knowledge. However, these efforts remain largely limited to \emph{static} images.
Developing a high-quality, multidisciplinary video benchmark presents greater challenges than those for text or image-based tasks due to the scarcity of suitable resources (\eg, textbooks or exam questions). This leaves the critical modality of videos and video-based expert-level reasoning significantly underexplored.
Recent work, MMWorld~\cite{he2024mmworld}, has made pioneering strides by incorporating videos across multiple disciplines. However, only a limited portion of its dataset (\mmworldrate) requires domain-specific expertise\footnote{
To estimate the proportion of MMWorld examples requiring domain expertise, we randomly sampled 200 instances from the human-annotated subset and engaged three annotators for evaluation. An example was classified as requiring domain expertise if at least one annotator marked it as such. 
}, and 76.4\% of the examples are generated by the GPT-4V model.
Moreover, most existing benchmarks provide only the ground-truth answer, restricting researchers' ability to conduct a fine-grained evaluation. 
To address this limitation, \ours includes expert-annotated reasoning rationales and relevant domain knowledge for each example, enabling a more nuanced assessment of expert-level reasoning.
\Cref{tab:comparison} further distinguishes the difference between \ours and existing multi-discipline benchmarks.

%% file: tables/data_comparison_table.tex
\begin{table}[!t]
\caption{Comparison between \ours and existing \multidis benchmarks for evaluating foundation models. In the ``QA Type'' column, ``MC'' denotes Multiple-Choice questions, ``Open'' denotes Open-ended questions, and ``T/F'' denotes True-False questions.}
\resizebox{\textwidth}{!}{%
\addtolength{\tabcolsep}{-0.45em}
\begin{tabular}{llp{5.6cm}>{\centering\arraybackslash}p{0.9cm}cc}
    \toprule
    \multirow{2}{*}{\textbf{Dataset}} & \multirow{2}{*}{\textbf{QA Type}} & \multirow{2}{*}{\textbf{Data Source}} & \multirow{2}{*}{\textbf{\begin{tabular}[c]{@{}c@{}}College\\Level?\end{tabular}}} & \multicolumn{2}{c}{\textbf{Detailed Solution}}\\ 
    & & & & \textbf{~~Rational?} & \textbf{Knowledge?}
    \\
    \midrule
    \multicolumn{5}{c}{\emph{\textbf{Text}}} \\
    \noalign{\vskip 1ex}
    MMLU~\cite{hendrycks2021mmlu} & MC & Exam, course, textbook & \cmark & \xmark & \xmark\\
    MMLU-Pro~\cite{wang2024mmlupro} & MC & Datasets $\rightarrow$ Human \& LLM augment & \cmark & \xmark & \xmark \\
    C-Eval~\cite{huang2023cevalmultilevelmultidisciplinechinese} & MC & Exam & \cmark & \xmark & \xmark \\
    SciEval~\cite{scieval} & MC, Open & Internet, datasets $\rightarrow$ LLM rewrite & \cmark & \xmark & \xmark \\
    TheoremQA~\cite{chen-etal-2023-theoremqa} & MC, T/F, Open & Internet, exam $\rightarrow$ Human rewrite& \cmark & \xmark & \cmark \\
    \multirow{2}{*}{SciKnowEval~\cite{feng2024sciknoweval}} & \multirow{2}{*}{MC, T/F, Open~} & Textbooks, database, other datasets $\rightarrow$ LLM rewrite & \multirow{2}{*}{\cmark} & \multirow{2}{*}{\xmark} & \multirow{2}{*}{\cmark} \\
    
    \midrule
    \multicolumn{5}{c}{\emph{\textbf{Text + Image}}} \\
    \noalign{\vskip 1ex}
    
    VisScience~\cite{jiang2024visscienceextensivebenchmarkevaluating} & MC, Open & Internet, exam, textbook & \xmark & \xmark & \xmark\\
    
    EXAMS-V~\cite{das2024examsvmultidisciplinemultilingualmultimodal} & MC & Exam & \xmark & \xmark & 
    \xmark\\
    
    ScienceQA~\cite{NEURIPS2022_11332b6b} & MC & Internet, course & \xmark &  \cmark & \xmark\\
    SceMQA~\cite{liang2024scemqascientificcollegeentrance} & MC, Open & Internet, exam & \xmark & \cmark & \xmark\\
    
    CharXiv~\cite{wang2024charxiv} & Open & arXiv paper $\rightarrow$ Human annotate & \cmark & \xmark & \xmark\\
    
    MMSci~\cite{li2024mmsci} & MC & Scientific paper $\rightarrow$ LLM generate & \cmark & \xmark & \xmark\\
    OlympicArena~\cite{huang2024olympicarenabenchmarkingmultidisciplinecognitive} & MC, T/F, Open & Olympic competitions & \cmark & \cmark & \xmark\\
    
    MMMU~\cite{yue2024mmmu} & MC, Open & Internet, exam, textbook & \cmark & 17.6\% & \xmark\\

    CMMMU~\cite{zhang2024cmmmu} & MC, T/F, Open & Internet, exam, textbook & \cmark & 2.1\% & \xmark\\
    
    MMMU-Pro~\cite{yue2024mmmupro} & MC & MMMU $\rightarrow$ Human \& LLM augment & \cmark & 15.4\% & \xmark\\

    \midrule
    \multicolumn{5}{c}{\emph{\textbf{Text + Video}}} \\
    \noalign{\vskip 1ex}
    MMWorld~\cite{he2024mmworld} & MC & Human experts (24\%) / LLM-gen (76\%) & \mmworldrate & \xmark & \xmark\\
    \noalign{\vskip 0.5ex}\hdashline\noalign{\vskip 0.5ex}
    \textbf{\ours (ours)} & MC, Open & Human experts annotate from scratch & \cmark & \cmark & \cmark \\ 
    \bottomrule
\end{tabular}
}
\label{tab:comparison}
\end{table}

%% file: main/3-dataset.tex
\section{\ours Benchmark}\label{sec:data}
We present \ours, a comprehensive evaluation benchmark that focuses on measuring progress on knowledge-intensive, expert-level reasoning in the video modality. 
\ours has the following key features:
(1) \textbf{Breadth of Domain Knowledge}:
We employ a textbook-guided QA annotation pipeline to ensure the wide coverage of domain knowledge within each subject (\S\ref{sec:data_collection}).
(2) \textbf{Depth of Expert-level Reasoning}:
Each example in \ours requires models to comprehend specialized-domain video context, applying expert knowledge and reasoning (\S\ref{sec:data_collection}).
(3) \textbf{True Visual Understanding}:
Recent studies~\cite{yue2024mmmupro, chen2024mmstar, zhang2024mathverse} have shown that visual content is unnecessary for many examples in current multimodal benchmarks.
To alleviate this issue, each example in \ours is carefully validated by human experts to confirm that video comprehension is required for accurate answering (\S\ref{sec:data_validation}). 
(4) \textbf{Support of Fine-grained Evaluation}:
We provide expert-annotated solutions and the requisite knowledge for each example (\S\ref{sec:data_collection}), enabling more comprehensive analysis for future research (\S\ref{sec:error_analysis}).
\Cref{fig:data_construction_pipeline} provides an overview of the three stages involved in constructing \ours, which is detailed in the following subsections.

\subsection{Preliminary Setup}
\input{figures_tex/data_pipeline}
We first discuss the preliminary setup for data construction. 

\paragraph{Subject Selection.} \label{sec:data_subject}
To ensure a broad and accurate representation of expert-level video understanding across diverse disciplines, we conduct a user study involving 133 college and graduate students for subject selection.
We ask them to curate two QA examples requiring expert-level video understanding in subjects relevant to their field of study, and provide feedback on their experiences during the curation process. 
Such a user study-guided approach helps us identify subjects within each discipline that may not be obvious from a top-down selection process. 
It also offers insights into the challenges of designing expert-level video examples, helping us design and refine the textbook-guided QA annotation process (detailed in \S\ref{sec:data_collection}). 
The authors manually analyze the collected examples and select \textbf{\nsubjects subjects} (as listed in \Cref{fig:overview}) across four disciplines that align best with our benchmark’s construction desiderata discussed earlier.

\paragraph{Expert Annotator Recruitment and Training.} 
For each subject, we assign at least two annotators with relevant expertise.
We include a total of 67 expert annotators (detailed biographies are presented in \Cref{app:annotator}), comprising 22 third- or fourth-year undergraduate students, 36 graduate students, and nine of the authors. All the annotators also participated in our initial user study.
Each annotator is required to finish a training session to learn the annotation protocol (detailed in \Cref{app:data_annotation_protocol}) before official annotation. 

\subsection{Textbook-Guided QA Example Annotation} \label{sec:data_collection}
Constructing a high-quality, expert-level, multi-disciplinary  benchmark for video-based tasks is more challenging than the ones for text- or image-based, as there is no existing resources (\eg, textbooks or exam questions) that can adapted from and each example has to be curated from scratch. 
Therefore, it is crucial to establish a structured approach that ensures the quality and comprehensiveness of the benchmark. 
We employ a textbook-guided example annotation pipeline designed to capture both the \emph{breadth of knowledge} and \emph{depth of reasoning}.
In brief, annotators first identify key concepts from the textbook and locate relevant videos that align with these concepts. The textbooks for each subject (listed in \Cref{app:textbook_selection}) are selected by expert annotators and are recognized as authoritative references in their respective fields. Annotators then curate QA examples and detailed solution rationales. We detail the annotation procedure as follows:

\paragraph{Concept-Driven CC-Licensed Video Collection.}
Annotators are instructed to first review each chapter of the textbook to identify key concepts that inherently require dynamic visual representation, such as experimental procedures in science or mechanical operations in engineering. 
They then search for related videos on YouTube having Creative Commons license\footnote{
The Creative Commons license 
enables reusers to distribute, remix, adapt, and build upon the material in any medium or format, so long as attribution is given to the creator.  
We use YouTube Data API v3 (\url{https://developers.google.com/youtube/v3}) to verify the license type. Existing video benchmarks typically utilize YouTube videos, yet do not confine their selections to content with CC licenses, introducing potential copyright concerns. We recognize that by restricting our selection to CC-licensed content, we are compelled to forgo coverage of certain subjects (\eg sports), where CC-licensed videos is scarce.
} 
that effectively illustrate the selected concept. 
To ensure the collected videos effectively challenge the model's visual reasoning capabilities, the video should be vision-intensive, requiring models to focus solely on visual information for comprehension. To this end, we ensure that audio tracks are excluded to eliminate potential shortcuts models might exploit through auditory cues; 
and the video should contain minimal on-screen text, as an overabundance of text may detract from the core visual understanding task. Consequently, videos such as lecture recordings, which typically include slides or text-based explanations that simplify the task of answering associated questions, are excluded.

\input{figures_tex/detailed_example}

\paragraph{QA Annotation.}
After identifying suitable videos, annotators 
are required to create two or three questions, either multiple-choice or open-ended.
Each question is designed to test the model's expert-level reasoning by applying domain-specific knowledge to interpret the video content and derive a solution.
Annotators are also required to specify the start and end timestamps of the video clip relevant to answering each question.
For annotating multi-choice question, the annotators are required to carefully craft the four distractor options to reflect common misconceptions or plausible alternatives, ensuring that models cannot easily eliminate incorrect options without reasoning over video content. Once the five options are finalized, the annotation interface randomly shuffles them.

\paragraph{Solution Rationale Annotation.}
For each annotated question, annotators must also provide detailed solution for the correct answers. As shown in \Cref{fig:detailed_example}, the solution comprises two key components: 
(1) \emph{relevant domain knowledge}, which includes a list of domain-specific concepts or keywords necessary for answering the question, with each concept linked to its corresponding Wikipedia page.
(2) \emph{reasoning rationale}, which details the step-by-step reasoning process to reach the correct answer. 
These solution annotations are critical for enhancing transparency in the evaluation process and facilitating future research focused on understanding model failure modes.

\subsection{Data Quality Control}\label{sec:data_validation}
We next discuss our methods to ensure high data quality.

\paragraph{Time-Based Annotation Compensation.} As discussed earlier, annotating examples for \ours can be particularly time-intensive, especially when there is limited availability of videos with Creative Commons licenses in the required subjects. 
To accommodate this and ensure a high-quality benchmark, we compensate annotators based on the time they spend rather than the number of examples completed, preventing them from rushing through tasks (See \Cref{app:payment} for annotation compensation details). On average, annotating one example takes 20 minutes and 17 seconds, while validation requires 4 minutes and 12 seconds.

\paragraph{Human Expert Validation.}
To ensure that the final dataset remains high-quality and meets expert-level standards without introducing unnecessary biases, each example in \ours undergoes expert review by one of the authors or top-performing annotators to verify the accuracy of its annotations. 
Recent studies~\cite{yue2024mmmupro, chen2024mmstar, zhang2024mathverse, shangguan2024tomato} have shown that visual content is unnecessary for many examples in current multimodal benchmarks.
To address this concern, each example in \ours is carefully validated by human experts to ensure that video comprehension is required for accurate answering. 
If an example is determined to be answerable solely through the textual components of the question, a single video frame, or if it contains annotation errors, evaluators first attempt to revise the example. 
If revision is not feasible, detailed feedback is provided to the original annotator, who then revises and submits it for a second iteration. 
A total of 523 examples were revised during the data validation process. Among them, 72 examples were still found to be misaligned with our design criteria and were excluded from the final benchmark. 
Overall, $1 - \frac{523}{3,000 + 72} = 83.0\%$ of the initial examples met our design criteria without requiring revisions, indicating the high quality of initial annotation.

\input{figures_tex/data_statistics}
\subsection{\ours Benchmark Analysis} \label{sec:benchmark-analysis}
\paragraph{Data Statistics.}
\Cref{tab:statistics} presents the key statistics of \ours. It consists of \nexample examples, which are randomly divided into two subsets: validation and test. 
The validation set contains \nval examples, and is intended for model development and validation. 
The test set, comprising the remaining \ntest examples, is strictly reserved for standard evaluation to prevent data contamination~\cite{jacovi-etal-2023-stop, deng-etal-2024-unveiling, glazer2024frontiermath}. To further promote fair benchmarking, the test set remains hidden. We are developing an online evaluation pipeline on a public platform, enabling researchers to benchmark their models and participate in a public leaderboard.

\paragraph{Human Performance.} 
To provide a rough but informative estimate of human-level performance on \ours, we randomly sampled 30 questions per discipline from the test set, resulting in a total of 120 questions for evaluation. 
Five participants—three graduate students specializing in biology, anesthesiology, and East-Asian literature, along with two of the authors—individually answered these questions. The evaluation proceeded in three phases:
(1) \textbf{Closed-book Setting}: In the first phase, participants had 3.5 hours to answer questions without access to external resources. The average accuracy across the four participants was 49.7\%. 
(2) \textbf{Open-book Setting}: In the second phase, participants were permitted to use external resources (\eg, internet and textbooks) to review answers they felt uncertain about. They were not informed of the correctness of their initial responses, and a 4-hour time limit was set. This open-book approach led to an increase in average accuracy to 86.8\%.
(3) \textbf{Oracle Setting}: Finally, participants were required to revise each incorrect answer based on ground-truth domain knowledge and self-sourced online resources. The average accuracy after this final revision was 95.3\%.

%% file: figures_tex/data_pipeline.tex
\begin{figure*}[!t]
 \centering
\includegraphics[width=\textwidth]{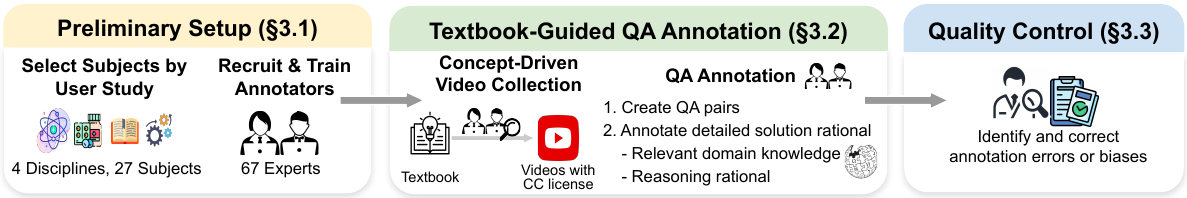}
 \caption{An overview of the \ours benchmark construction pipeline.}
 \label{fig:data_construction_pipeline}
\end{figure*}

%% file: figures_tex/detailed_example.tex
\begin{figure}[!t]
    \centering
    \includegraphics[width=\textwidth]{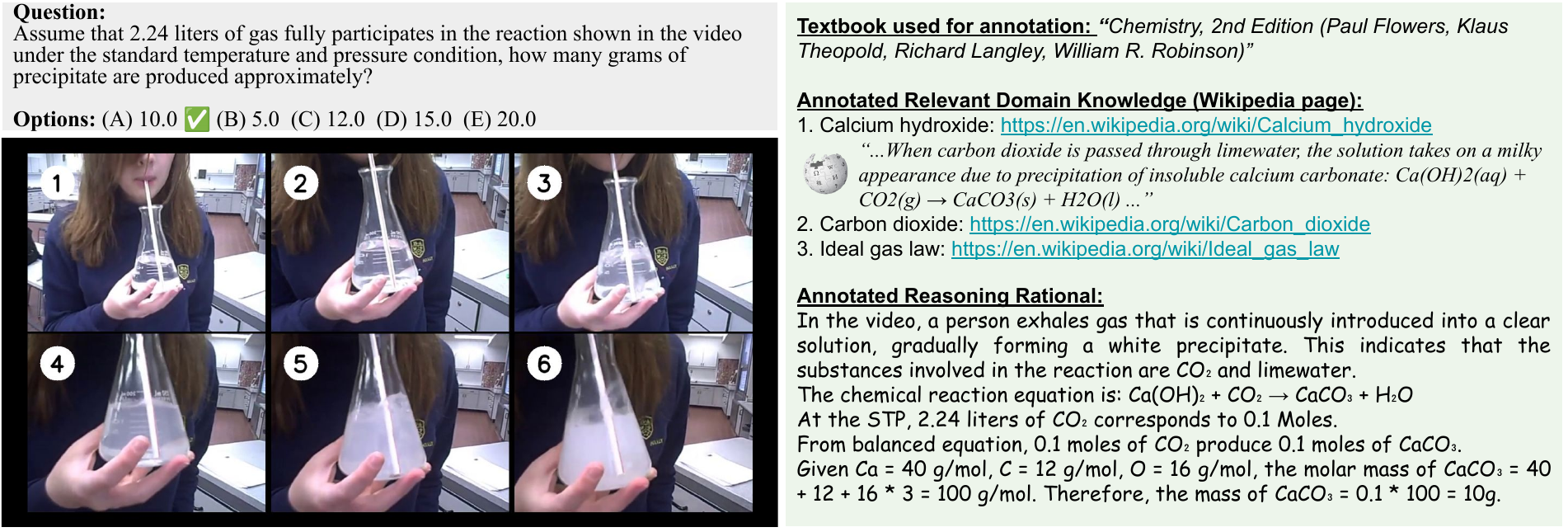}
    \captionof{figure}{
    A dataset example from \ours with the discipline of chemistry. Each example in \ours includes expert annotation of relevant domain knowledge and step-by-step reasoning rational. 
    }
    \label{fig:detailed_example}
\end{figure}

%% file: figures_tex/data_statistics.tex
\begin{table}[!t]
\centering
\footnotesize
\centering
 \renewcommand\tabcolsep{1.0pt}
\caption{Key statistics of the \ours benchmark.}
 \begin{tabular}{lr}
 \toprule
 \textbf{Statistics} & \textbf{Value} \\
 \midrule
  Total Questions & \nexample \\
  \quad Validation Set & \nval \\
  \quad Test Set & \ntest \\
  
\midrule
 Unique Videos  & \nvideo \\
 Video Length (Seconds, \texttt{avg/max})  & 51.4 / 228 \\
 \midrule
Number of Disciplines & 4 \\
Number of Subjects & \nsubjects \\

\midrule 
Multiple Choice Questions  & 1,858\\
\quad Question Length (\texttt{avg/max})  & 16.8 / 70 \\
\quad Single Choice Length (\texttt{avg/max})  & 7.6 / 42 \\
\quad Number of Choices per Question & 5 \\

\noalign{\vskip 0.5ex}\hdashline\noalign{\vskip 0.5ex}

Open-ended Questions  & 1,142 \\
\quad Question Length (\texttt{avg/max})  & 16.4 / 39 \\
\quad Ground-truth Answer Length (\texttt{avg/max})  & 1.5 / 7 \\

\midrule
Number of Required \textbf{Knowledge} per Question (\texttt{avg/max}) & 4.3 / 7\\
\textbf{Solution Rationale} Length (\texttt{avg/max})  & 56.6 / 193 \\
\midrule
Total Number of Unique Knowledge (\ie, Wikipedia pages) & 4,770 \\
 \bottomrule
 \end{tabular}
\label{tab:statistics}
\end{table}

%% file: main/4-experiment.tex
\section{Experiments}\label{sec:main-exp}
This section discusses the experiment setup and our key findings.

\subsection{Experiment Setup}
\paragraph{Evaluated Multimodal Foundation Models.} \label{sec:model}
To establish a comprehensive understanding of the challenges posed by \ours and provide reference points for future research, we evaluate a broad range of frontier multimodal foundation models that support \emph{video} or \emph{multiple images} as input. 
Specifically, we evaluate \textbf{16 series of open-source models}, including 
InternVL-2 \& 2.5~\cite{chen2023internvl, chen2024fargpt4vclosinggap}, 
Qwen2-VL~\cite{wang2024qwen2vl, yang2024qwen2}, 
LLaVA-NeXT~\cite{liu2024llavanext},
Pixtral~\cite{pixtral},
DeepSeek-VL2~\cite{wu2024deepseekvl2mixtureofexpertsvisionlanguagemodels},
H2OVL Mississippi~\cite{galib2024h2ovlmississippivisionlanguagemodels},
Idefics2~\cite{laurençon2024mattersbuildingvisionlanguagemodels},
Aria~\cite{li2025ariaopenmultimodalnative}, 
LLaVA-NeXT-Video~\cite{li2024llava},
LLaVA-OneVision~\cite{li2024llavaonevisioneasyvisualtask},
Llama-3.2-Vision~\cite{dubey2024llama3herdmodels},
Phi-3.5-Vision~\cite{abdin2024phi3},
InternVideo2~\cite{wang2024internvideo2},
and VideoLLaMA2 \& 2.1~\cite{damonlpsg2024videollama2}.
We also evaluate \textbf{eight series of proprietary models}, including 
OpenAI o1~\cite{Contributors2024OpenAIOS} and GPT-4o~\cite{openai2024gpt4o}, 
Gemini-1.5 \& 2 and Gemini-Thinking~\cite{geminiteam2024gemini}, 
GLM-4V-Plus~\cite{glm2024chatglmfamilylargelanguage},
Grok-2-Vision~\cite{grok},
and Claude-3.5~\cite{claude3}.
For open-source models, we prioritize the vLLM pipeline~\cite{kwon2023efficient} for model inference; otherwise, we use the Transformers pipeline~\cite{wolf-etal-2020-transformers}.
We use the official API service for proprietary models.
For models without native video support, following VideoMME~\cite{fu2024videomme}, we provide visual input using the maximum number of images that fits within the model’s context window.
\S\ref{app:model_info} details the parameter settings and model configurations.
We evaluate the models with both \textbf{Direct Answer} and \textbf{Chain-of-Thought} prompts (presented in \cref{app:cot_prompt}), which is adapted from the versions used in MMMU-Pro~\cite{yue2024mmmupro}. 

\paragraph{Accuracy Evaluation.}\label{sec:evaluation-system}
We use accuracy as the primary metric to evaluate model performance on \ours.
Following recent benchmarks for foundation model evaluation~\cite{wang2024charxiv, lu2024mathvista, he2024mmworld}, we employ GPT-4o to assess accuracy. Specifically, given a question, its ground truth answer, and the model's response, GPT-4o is instructed to extract the final answer from the model response and determine its correctness. The evaluation prompts for both multiple-choice and open-ended questions are presented in \Cref{app:acc_evaluation_prompt}.

\input{main/4-1-main_results}

%% file: main/4-1-main_results.tex
\subsection{Main Findings}
\Cref{tab:main_results} presents the evaluated models' CoT performance on the \ours benchmark, while \Cref{fig:cot_compare_main} illustrates a comparison between the model performance in CoT reasoning and direct answering. Our key findings are as follows:
\input{tables/main_results}

\paragraph{\ours presents substantial challenges for current multimodal foundation models.}
Even the top-performing model falls well short of human expert performance. For instance, GPT-4o achieves \gpt accuracy with CoT prompting, significantly lower than the 86.8\% accuracy achieved by human experts in an open-book setting.
Notably, while GPT-4o has narrowed the performance gap with human experts in text-based expert-level reasoning on MMLU (88.7\% vs 89.8\% \cite{hendrycks2021mmlu}) and image-based expert-level reasoning on MMMU (69.1\% vs 82.6\% \cite{yue2024mmmu}), the gap remains large on \ours.
This disparity underscores \ours's critical role in advancing and evaluating multimodal foundation models' capabilities in video-based expert reasoning across specialized domains.

\paragraph{Performance of open-sourced models.}
As for open-source multimodal foundation models, they still lag behind the proprietary models. However, the Qwen2-VL-72B and DeepSeek-VL2 models have achieved performance levels that exceed human benchmarks in closed-book settings and are approaching the performance of leading proprietary models. These advancements highlight the significant progress being made in the development of open-source models.

\paragraph{CoT reasoning generally improves model performance compared to directly outputting the answer.} 
However, the degree of improvement varies across different foundation models.
For instance, Claude 3.5 Sonnet demonstrated a remarkable enhancement, achieving a notable performance gain of 11.0\%, as corroborated by the findings in MMMU-Pro~\cite{yue2024mmmupro}.  \input{figures_tex/cot_do_comparison}
Conversely, models like GPT-4o exhibited only marginal improvements.
These results indicate that the impact of CoT reasoning is not uniformly beneficial across all models on \ours.

\paragraph{System-2 thinking demonstrates effectiveness.} 
Models capable of System-2 thinking and employing long CoT demonstrate significant performance advantages. Notably, the o1 and Gemini 2.0 Flash Thinking models achieved the top two results on \ours, illustrating that increasing test-time compute and applying long CoT can significantly enhance model performance in expert-level video reasoning tasks. 
These results highlight the potential of developing open-source models designed to facilitate and advance System-2 thinking capabilities.

\subsection{Qualitative Analysis}\label{sec:error_analysis}
To gain a deeper understanding of the capabilities and limitations of frontier models on \ours, we perform comprehensive case studies and error analysis by humans. 
The inclusion of expert-annotated reasoning rationales and domain knowledge for each example in \ours facilitate a more effective analysis compared to datasets that provide only answers.
We focus on four top-performing models, GPT-4o, Qwen2-VL-72B, Llama-3.2-90B-Vision, and DeepSeek-VL2, for human evaluation.
From the \ours validation set, we randomly sample 50 error cases for each model. 
These cases are analyzed by the authors using ground-truth features (\ie expert-annotated reasoning rationales and required domain knowledge) as references. We identify following \nerror primary errors:

\noindent\textbf{Visual Perception Error (18\%):} 
The model fails to accurately interpret spatial, temporal, or semantic aspects of visual information within a video. Additionally, it might ``hallucinate'', detecting objects or events that are not actually present in the video. \autoref{fig:main_error_analysis} (left) is a typical related instance where the model fails to correctly perceive the traversal order of the binary tree. Similarly, \autoref{fig:visual_perception_error_2} shows that the model mistakenly identifies the device shell in the video as water, leading to completely wrong reasoning about the device's function.

\noindent\textbf{Misuse or Lack Domain Knowledge in Visual Perception (20\%):} 
The model fails to apply the domain-specific expertise required to accurately interpret specialized concepts or elements within the video. For example, in a medical video, it may identify objects but fail to recognize their technical terms or misunderstand their importance within the procedure being demonstrated. Moreover, as shown in \autoref{fig:visual_perception_misuse_1}, the model correctly perceives the ascending numbers (array indices), but misuses its pretrained knowledge and misidentifies them as the numbers to be sorted. It leads to the wrong conclusion that the video demonstrates a sorting algorithm. This limitation underscores a gap in the model’s ability to integrate domain knowledge with visual perception effectively. 

\noindent\textbf{Misuse or Lack Domain Knowledge in Reasoning (27\%):}
The model fails to effectively recall and apply domain knowledge during its reasoning processes. For instance, when addressing questions over chemistry videos, it may fail to correctly apply relevant chemical equations, leading to errors in computing the reaction mass. A notable example is \autoref{fig:main_error_analysis} (right), where the model misuses the domain knowledge that bats often live in unsanitary environments and makes the wrong inference that poor hygiene conditions are the cause of virus outbreaks. Besides, in \autoref{fig:knowledge_reasoning_3}, the model lacks the domain knowledge about relevant chemical equations, so that it cannot correctly answer the question.
This limitation underscores the model's inability to integrate domain knowledge into its reasoning processes effectively.

\noindent\textbf{Heavy Reliance on Textual Information (20\%):}
The model predominantly depends on textual information for problem-solving, especially when addressing multiple-choice questions, as it evaluates each option individually without leveraging the actual video content. For instance, \autoref{fig:text_error_1} shows the model ignores the video information about the reason of the disease and overly focuses on the textual question. Similar limitations have been observed in other multimodal benchmarks~\cite{fu2024videomme, yue2024mmmu}. This gap suggests future work in enhancing multimodal reasoning by more effectively incorporating non-textual content into the reasoning process.

\noindent\textbf{Logical Reasoning Error (6\%):}
The model exhibits inconsistencies between its reasoning process and final answer, leading to self-contradiction. As depicted in \autoref{fig:logical_error_1}, the analysis of one specific option contradicts with the other reasoning steps, which is a typical self-contradiction logical error.

\noindent\textbf{Other Error (9\%):}
This includes other errors, such as refusing to answer a question due to insufficient context or safety concerns, generating a response that exceeds the output limit, generating repetitive information, or making incorrect math computation.

\input{figures_tex/main_error_analysis}

%% file: tables/main_results.tex
\begin{table*}[!t]
\caption{
Accuracy of evaluated foundation models on the \ours validation and test sets using CoT prompts. 
Model performance is ranked based on overall results on the test set.
$^*$: For o1, as the API access for its multimodal version has not been granted, we randomly sampled 100 examples from the validation set and 200 examples (50 for each core discipline) from the test set. The model’s performance was manually evaluated on Jan 10, 2025, using CoT prompts on ChatGPT platform.
}
\label{tab:main_results}
\centering
\renewcommand\arraystretch{1.1} 
\resizebox{1\linewidth}{!}{%
\footnotesize
\begin{tabular}{ll*{4}{>{\centering\arraybackslash}p{1.4cm}}*{2}{>{\centering\arraybackslash}p{1.1cm}}}
\toprule[1pt]
 & \multirow{3}{*}{\textbf{Release}} &\multicolumn{4}{c}{\textbf{Test Set}} 
& \multirow{3}{*}{\textbf{\begin{tabular}[c]{@{}c@{}}Avg. \\  Validation\end{tabular}} } & \multirow{3}{*}{\textbf{\textbf{\begin{tabular}[c]{@{}c@{}}Avg. \\  Test\end{tabular}} }} \\
\cline{3-6}\noalign{\vskip 1ex}
 & & \textbf{Science} & \textbf{Healthcare} & \textbf{\begin{tabular}[c]{@{}c@{}}Human. \&\\  Social Sci.\end{tabular}} & \textbf{Engineering} \\
\midrule[0.7pt]
\multicolumn{8}{c}{\textbf{\textit{Human Performance}}} \\
\noalign{\vskip 1ex}
Human Oracle & & 95.3 & 93.3 & 96.0 & 96.7 & \multicolumn{2}{c}{95.3}\\
Human Open-book & & 86.7 & 84.7 & 92.7 & 83.3 & \multicolumn{2}{c}{86.8}\\
Human Closed-book & & 54.7 & 42.7 & 44.7 & 56.7 & \multicolumn{2}{c}{49.7}\\

\midrule
\multicolumn{8}{c}{\textbf{\textit{Proprietary Models}}} \\
\noalign{\vskip 1ex}
o1$^*$ & 2024-12 &     \cellcolor{red!35}{80.0} &        \cellcolor{red!35}78.0 &                     \cellcolor{red!35}76.0 &                \cellcolor{red!35}74.0 &     \cellcolor{red!35}79.0 &      \cellcolor{red!35}77.0 \\
Gemini 2.0 Flash Thinking & 2024-12 &     \cellcolor{red!5}{69.3} &        71.2 &                     \cellcolor{red!20}73.4 &                \cellcolor{red!20}67.3 &     \cellcolor{red!20}69.1 &      \cellcolor{red!20}69.5 \\
GPT-4o &  2024-08 &     67.2 &        \cellcolor{red!5}71.8 &                     \cellcolor{red!5}72.0 &                61.6 &     \cellcolor{red!5}67.4 &      \cellcolor{red!5}66.7 \\
Gemini 2.0 Flash & 2024-12 &     \cellcolor{red!20}{70.8} &        62.7 &                     71.6 &                63.0 &     65.9 &      66.5 \\
Gemini 1.5 Pro & 2024-09 &     67.2 &        68.1 &                     67.0 &                62.8 &     65.4 &      65.8 \\
Claude 3.5 Sonnet & 2024-10 &     60.5 &        64.0 &                     70.9 &                \cellcolor{red!5}64.5 &     65.2 &      64.1 \\
Grok-2-Vision & 2024-12 &     60.6 &        \cellcolor{red!20}72.5 &                     \cellcolor{red!5}72.0 &                57.4 &     62.7 &      63.4 \\
GPT-4o-mini & 2024-07 &     60.3 &        60.9 &                     70.6 &                59.3 &     61.6 &      61.5 \\
Gemini 1.5 Flash & 2024-09 &     56.8 &        57.3 &                     66.3 &                58.2 &     58.8 &      58.8 \\
GLM-4V-Plus & 2025-01 &     52.2 &        57.3 &                     64.9 &                55.4 &     56.2 &      56.2 \\

\midrule
\multicolumn{8}{c}{\textbf{\textit{Open-sourced Models}}} \\
\noalign{\vskip 1ex}
Qwen2-VL-72B & 2024-09 &     48.0 &        53.6 &                     61.7 &                53.9 &     53.0 &      53.2 \\
DeepSeek-VL2 & 2024-12 &     50.3 &        53.4 &                     58.9 &                48.6 &     52.1 &      51.5 \\
InternVL2.5-38B & 2024-11 &     50.3 &        45.6 &                     52.8 &                52.8 &     50.5 &      50.7 \\
Aria & 2024-11 &     46.8 &        43.3 &                     61.0 &                49.9 &     49.3 &      49.3 \\
Llama-3.2-90B-Vision & 2024-09 &     46.5 &        43.5 &                     53.9 &                48.1 &     47.1 &      47.6 \\
DeepSeek-VL2-Small & 2024-12 &     47.5 &        48.7 &                     47.5 &                45.1 &     46.9 &      46.9 \\
Qwen2-VL-7B-Instruct & 2024-08 &     43.6 &        42.5 &                     43.6 &                41.2 &     42.1 &      42.5 \\
InternVL2.5-8B & 2024-11 &     39.2 &        36.8 &                     47.2 &                42.3 &     41.1 &      41.0 \\
VideoLLaMA2.1-7B & 2024-10 &     35.3 &        38.9 &                     45.4 &                41.6 &     39.5 &      39.8 \\
Llama-3.2-11B-Vision & 2024-09 &     40.5 &        39.4 &                     44.0 &                35.7 &     38.9 &      39.0 \\
Phi-3.5-Vision & 2024-08 &     38.3 &        29.5 &                     45.4 &                41.1 &     38.1 &      38.7 \\
LLaVA-OneVision-7B & 2024-09 &     34.3 &        38.6 &                     40.8 &                38.8 &     37.9 &      37.7 \\
Qwen2-VL-2B & 2024-08 &     32.6 &        40.9 &                     40.4 &                35.7 &     36.5 &      36.5 \\
InternVL2-8B & 2024-06 &     36.7 &        32.9 &                     36.9 &                37.2 &     36.3 &      36.2 \\
Idefics3-8B & 2024-08 &     37.0 &        35.5 &                     44.0 &                31.2 &     35.3 &      35.6 \\
VideoLLaMA2-7B & 2024-06 &     32.3 &        27.7 &                     44.3 &                35.7 &     34.4 &      34.4 \\
DeepSeek-VL2-Tiny & 2024-12 &     34.3 &        33.4 &                     35.8 &                30.1 &     33.0 &      32.8 \\
Pixtral-12B & 2024-09 &     36.1 &        24.6 &                     37.9 &                30.8 &     32.3 &      32.2 \\
LLaVA-NeXT-Video-34B & 2024-06 &     31.8 &        24.6 &                     35.8 &                30.3 &     30.5 &      30.4 \\
InternVideo2-8B & 2024-08 &     29.6 &        31.1 &                     37.2 &                26.5 &     29.9 &      29.9 \\
H2OVL Mississippi-2B & 2024-10 &     29.1 &        29.5 &                     29.4 &                28.0 &     29.1 &      28.8 \\
LLaVA-NeXT-Video-7B & 2024-06 &     27.0 &        31.1 &                     27.3 &                29.5 &     28.6 &      28.7 \\
\bottomrule[1pt]
\end{tabular}
}
\end{table*}
\vspace{10pt}

%% file: figures_tex/cot_do_comparison.tex
\begin{wrapfigure}{r}{0.48\linewidth}
    \centering
    \includegraphics[width=\linewidth]{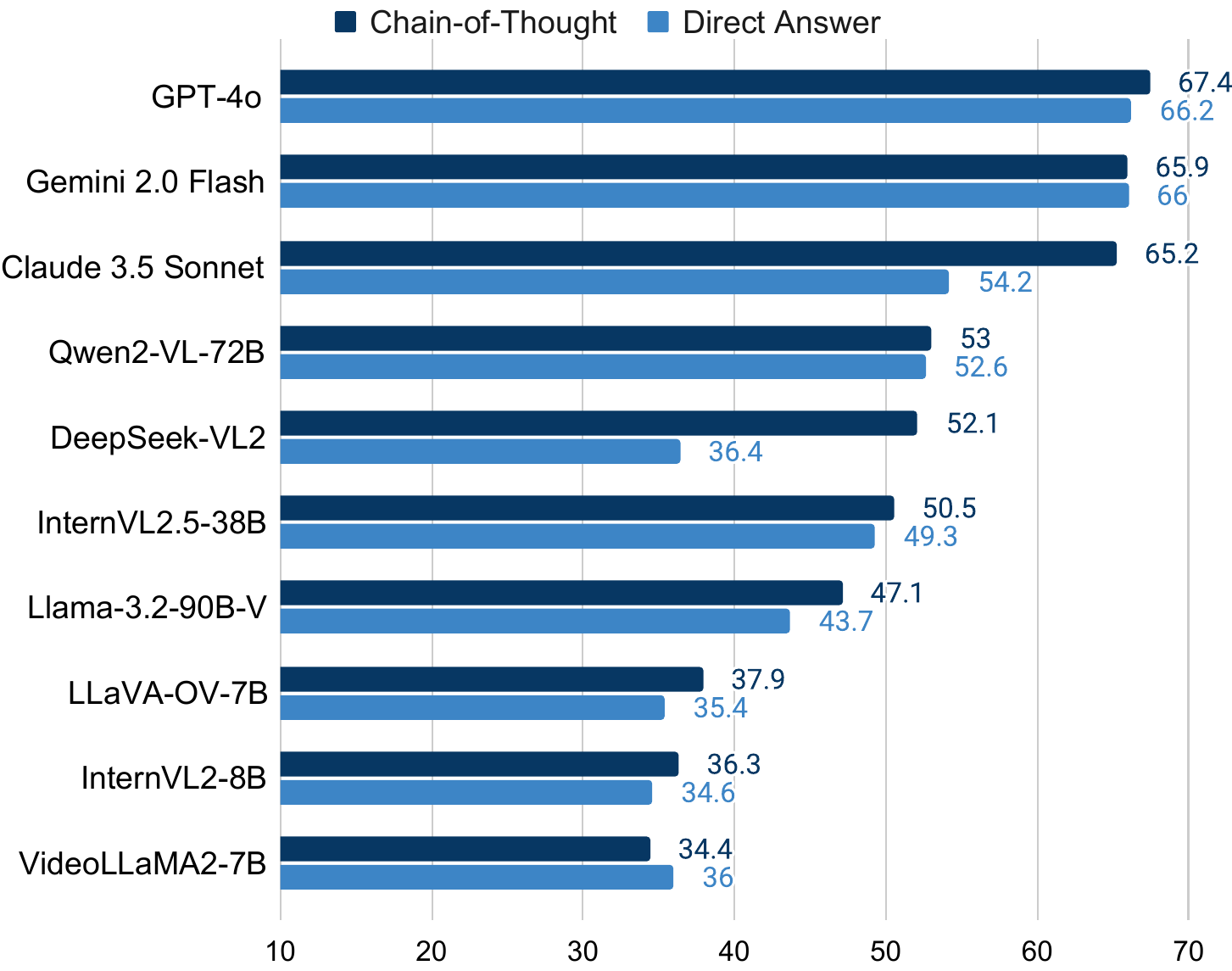}
    \caption{Comparison of model performance between CoT and direct answering on the validation set. The full results are provided in \S\ref{app:cot_compare}.}
    \label{fig:cot_compare_main}
    \vspace{-20pt}
\end{wrapfigure}

%% file: figures_tex/main_error_analysis.tex
\begin{figure}[!t]
    \centering
    \includegraphics[width=\textwidth]{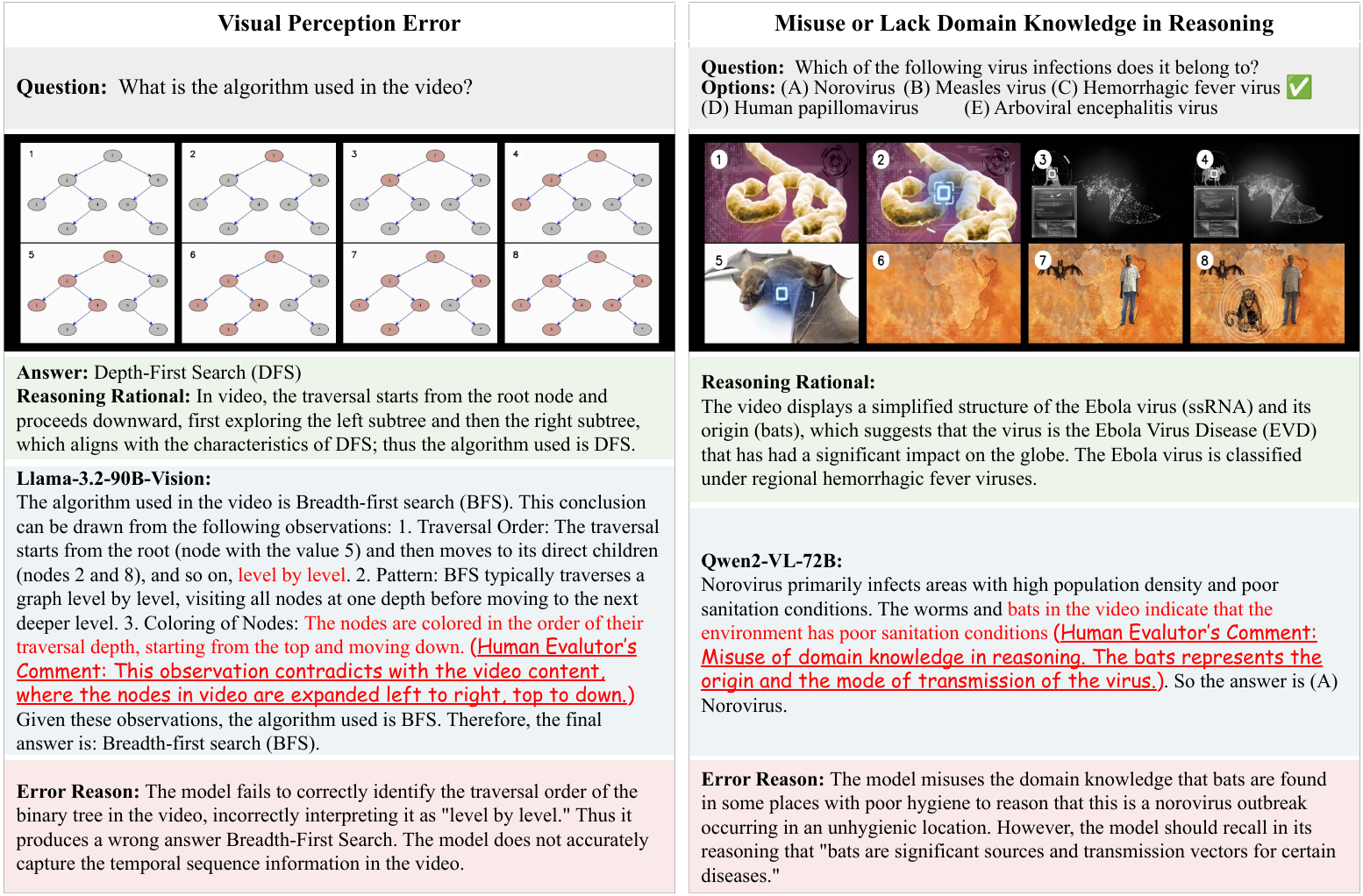}
    \captionof{figure}{
    Illustrations of visual perception error and misuse or lack domain knowledge in reasoning.
    }
    \label{fig:main_error_analysis}
\end{figure}

%% file: main/5-conclusion.tex
\section{Conclusion}
We introduce \ours, a high-quality, multi-disciplinary benchmark designed to assess the expert-level, knowledge-intensive reasoning capabilities of multimodal foundation models on specialized-domain videos. Each example in \ours is annotated by human experts from scratch.
We employ a textbook-guided example annotation pipeline designed to capture both the breadth of knowledge and depth of reasoning.
In our evaluation of \nmodel frontier multimodal foundation models, we find that while the latest o1 model achieves the highest performance among all tested models—approaching human expert-level proficiency—a notable performance gap remains between other models and human experts. 
Additionally, models employing CoT reasoning consistently outperform those that generate final answers directly.
Through comprehensive error analysis and case studies, we identify persistent challenges of \ours, offering valuable insights for advancing foundation models' capabilities to achieve expert-level video understanding in specialized domains.

\clearpage
\section*{Author Contribution}
The author contributions are summarized below:

\begin{itemize}[itemsep=1.6pt,leftmargin=12pt,topsep=1.6pt]
    \item {\bf Project Lead}: Yilun Zhao
    \item {\bf Project Conception}: Yilun Zhao, Lujing Xie, Yitao Long, Zhiyuan Hu, Zhenwen Liang, Xiangru Tang, Yixin Liu, Chen Zhao, Arman Cohan
    \item {\bf User Study}: Every author
    \item {\bf Data Annotation Protocol Development}: Yilun Zhao, Lujing Xie, Chengye Wang
    \item {\bf Data Annotation Task Management}: Lujing Xie, Haowei Zhang
    \item {\bf Data Annotation}: Lujing Xie, Haowei Zhang, Tongyan Hu, Weiyuan Chen, Junyang Song, Zhijian Xu, Weifeng Pan, Guo Gan, Yitao Long
    \item {\bf Data Validation}: Lujing Xie, Haowei Zhang, Tongyan Hu, Weiyuan Chen, Yilun Zhao, Junyang Song
    \item {\bf Data Annotation Expense}: Yilun Zhao
    \item {\bf Codebases and Results}: Yilun Zhao, Guo Gan
    \item {\bf Error Analysis and Case Study}: Haowei Zhang, Lujing Xie, Yilun Zhao, Weiyuan Chen
    \item {\bf Manuscript Writing}: Yilun Zhao, Haowei Zhang, Arman Cohan
    \item {\bf Manuscript Editing}: Every author
\end{itemize}

%% file: appendix/A-dataset_construction.tex
\clearpage
\section{\ours Preliminary Setup}
\subsection{Annotator Biography}\label{app:annotator}
The detailed biographies of the annotators involved in \ours construction are presented in \autoref{tab:annotator_bib}. All annotators are from universities ranked in the Top 500 of the 2024 QS Global Rankings\footnote{\url{https://www.topuniversities.com/world-university-rankings}} and are fluent in English.
\input{appendix/tables/A2-annotator_info}

\clearpage
\subsection{Textbook for Each Subject}\label{app:textbook_selection}
As discussed in \Cref{sec:data_collection}, we design a textbook-guided example annotation pipeline to encompass both the \emph{breadth of knowledge} and the \emph{depth of reasoning}.
The textbooks used for each subject are detailed in the following tables. 
They are selected by expert annotators and are recognized as authoritative references in their respective fields. 

\input{appendix/tables/A3-textbook}

\clearpage
\input{appendix/A-1-annotation_protocol}
\clearpage

%% file: appendix/tables/A2-annotator_info.tex
\begin{table*}[h]
\centering
\renewcommand{\arraystretch}{1.05}
\resizebox{\textwidth}{!}{%
\begin{tabular}{llllcc}
\toprule
\textbf{ID} & \textbf{Year} & \textbf{Major} & \textbf{Assigned Subject(s)} & \textbf{Author?} & \textbf{Validator?}\\
\midrule
1 & 1st year Master & Biomedical Engineering & Biomedical Engineering & \xmark & \xmark \\
 & & & Computer Science & & \\
 & & & Electrical Engineering & & \\
2 & 1st year Master & Bioinformatics & Biomedical Engineering & \xmark & \xmark \\
3 & 1st year Master & Biological Engineering & Biomedical Engineering & \xmark & \xmark \\
4 & 2nd year Master & Biomedical Engineering & Biomedical Engineering & \xmark & \xmark \\
 & & & Electronics and Communication & & \\
5 & 5th year PhD & Agricultural and Biosystems Engineering & Biomedical Engineering & \xmark & \xmark \\
6 & 2nd year Master & Architecture & Civil Engineering & \xmark & \xmark \\
7 & 3rd year PhD & Civil Engineering & Civil Engineering & \xmark & \xmark \\
 & & & Mechanical Engineering & & \\
8 & -- & -- & -- & \cmark & \cmark \\
9 & 3rd year Undergraduate & Electrical Engineering & Computer Science & \xmark & \xmark \\
 & & & Electrical Engineering & & \\
10 & 2nd year Master & Electrical Engineering & Computer Science & \xmark & \xmark \\
 & & & Electronics and Communication & & \\
11 & 2nd year Master & Electrical Engineering & Computer Science & \xmark & \xmark \\
 & & & Mechanical Engineering & & \\
12 & 3rd year Undergraduate & Software Engineering & Computer Science & \xmark & \xmark \\
13 & 2nd year Master & Computer Science & Computer Science & \xmark & \xmark \\
14 & -- & -- & -- & \cmark & \xmark \\
 & & & Electrical Engineering & & \\
15 & 1st year PhD & Electrical Engineering & Computer Science & \xmark & \xmark \\
 & & & Electronics and Communication & & \\
16 & 1st year PhD & Electrical Engineering & Electrical Engineering & \xmark & \xmark \\
17 & -- & -- & -- & \cmark & \cmark \\
18 & 1st year Master & Electrical Engineering & Electrical Engineering & \xmark & \xmark \\
 & & & Mechanical Engineering & & \\
19 & 1st year PhD & Electrical Engineering & Electronics and Communication & \xmark & \xmark \\
20 & 3rd year PhD & Food Science & Mechanics & \xmark & \xmark \\
21 & 4th year PhD & Materials Science & Materials Science & \xmark & \xmark \\
22 & 4th year Undergraduate & Aerospace Engineering & Materials Science & \xmark & \xmark \\
 & & & Mechanical Engineering & & \\
23 & 4th year Undergraduate & Mechanical Engineering & Materials Science & \xmark & \cmark \\
 & & & Mechanical Engineering & & \\
24 & 2nd year PhD & Mechanical Engineering & Mechanical Engineering & \xmark & \xmark \\
25 & 1st year PhD & Mechanical Engineering & Mechanical Engineering & \xmark & \xmark \\
26 & 1st year Master & Medicine & Basic Medicine & \xmark & \xmark \\
 & & & Clinical Medicine & & \\
27 & 1st year Master & Radiology & Basic Medicine & \xmark & \xmark \\
 & & & Clinical Medicine & & \\
28 & 1st year Master & Dentistry & Basic Medicine & \xmark & \xmark \\
 & & & Dentistry & & \\
29 & 1st year PhD & Nursing & Basic Medicine & \xmark & \xmark \\
 & & & Pharmacy & & \\
30 & 3rd year Undergraduate & Epidemiology & Basic Medicine & \xmark & \xmark \\
 & & & Preventive Medicine & & \\
31 & 3rd year Undergraduate & Medicine & Clinical Medicine & \xmark & \xmark \\
32 & -- & -- & -- & \cmark & \cmark \\
33 & 2nd year PhD & Medicine & Clinical Medicine & \xmark & \xmark \\
 & & & Pharmacy & & \\
 \bottomrule
\end{tabular}
}

\caption{
Biographies of 73 annotators involved in \ours construction (Author biographies are hidden to protect identity confidentiality).
}
\label{tab:annotator_bib}
\end{table*}

\begin{table*}[!t]
\centering
\renewcommand{\arraystretch}{1.05}
\resizebox{\textwidth}{!}{%
\begin{tabular}{llllcc}
\toprule
\textbf{ID} & \textbf{Year} & \textbf{Major} & \textbf{Assigned Subject(s)} & \textbf{Author?} & \textbf{Validator?}\\
\midrule
34 & 4th year PhD & Dentistry & Dentistry & \xmark & \xmark\\
35 & 3rd year Undergraduate & Dentistry & Dentistry & \xmark & \xmark \\
36 & 4th year PhD & Dentistry & Dentistry & \xmark & \xmark \\
37 & 1st year PhD & Public Health & Pharmacy & \xmark & \xmark \\
 & & & Preventive Medicine & & \\
38 & 4th year Undergraduate & Pharmacy & Pharmacy & \xmark & \xmark \\
39 & 3rd year PhD & East Asian Studies & Art & \xmark & \xmark \\
40 & 4th year PhD & Literature & Art & \xmark & \xmark \\
 & & & History & & \\
 & & & Literature & & \\
41 & -- & -- & -- & \cmark & \xmark \\
 & & & History & &\\
42 & 1st year PhD & Economics & Economics & \xmark & \xmark \\
43 & 4th year Undergraduate & Accounting & Economics & \xmark & \xmark \\
 & & & Law & & \\
44 & 4th year PhD & Finance & Economics & \xmark & \xmark \\
45 & 3rd year PhD & Public Administration & Law & \xmark & \xmark\\
 & & & Management & & \\
46 & 1st year Master & Literature & Literature & \xmark & \xmark\\
47 & 5th year PhD & Linguistics & Literature & \xmark & \xmark \\
48 & 3rd year Undergraduate & Public Administration & Management & \xmark & \xmark \\
49 & 5th year PhD & Astronomy & Astronomy & \xmark & \xmark \\
50 & -- & -- & -- & \cmark & \cmark \\
51 & 2nd year Master & Astronomy & Astronomy & \xmark & \xmark \\
52 & -- & -- & -- & \cmark & \xmark \\
 & & & Geography & & \\
53 & 3rd year PhD & Biology & Biology & \xmark & \xmark \\
54 & 1st year PhD & Biology & Biology & \xmark & \xmark \\
 & & & Neurobiology & & \\
55 & 3rd year PhD & Marine Biology & Biology & \xmark & \xmark \\
 & & & Chemistry & & \\
56 & -- & -- & -- & \cmark & \xmark \\
57 & 1st year PhD & Chemistry & Chemistry & \xmark & \xmark \\
58 & 3rd year Undergraduate & Chemistry & Chemistry & \xmark & \xmark \\
59 & 1st year PhD & Physics & Electromagnetism & \xmark & \xmark\\
60 & 4th year Undergraduate & Physics & Electromagnetism & \xmark & \xmark \\
 & & & Thermodynamics & & \\
61 & 4th year PhD & Physics & Electromagnetism & \xmark & \xmark \\
62 & 1st year PhD & Physics & Electromagnetism & \xmark & \xmark \\
 & & & Mechanics & & \\
 & & & Thermodynamics & & \\
63 & 1st year Master & Physics & Thermodynamics & \xmark & \xmark\\
 & & & Electromagnetism & & \\
64 & 3rd year Undergraduate & Agricultural and Environmental Sciences & Geography & \xmark & \xmark \\
65 & 4th year PhD & Physics & Thermodynamics & \xmark & \xmark \\
 & & & Mechanics & & \\
 & & & Modern Physics & & \\
66 & 1st year PhD & Physics & Mechanics & \xmark & \xmark \\
67 & 3rd year PhD & Physics & Mechanics & \xmark & \xmark \\
68 & 4th year PhD & Physics & Modern Physics & \xmark & \xmark \\
69 & 3rd year Undergraduate & Neurobiology & Neurobiology & \xmark & \xmark \\
70 & 1st year PhD & Neurobiology & Neurobiology & \xmark & \xmark \\
71 & -- & -- & -- & \cmark & \cmark \\
72 & 3rd year Undergraduate & Biology & Neurobiology & \xmark & \xmark\\
73 & 1st year Master & Biology & Neurobiology & \xmark & \xmark\\
\bottomrule
\end{tabular}
}

\caption{
Biographies of 73 annotators involved in \ours construction (Author biographies are hidden to protect identity confidentiality).
}
\end{table*}

%% file: appendix/tables/A3-textbook.tex
\begin{table*}[h]
\centering
\small
\renewcommand{\arraystretch}{1.05}
\begin{tabular}{p{2.5cm}p{10cm}}
\toprule
\textbf{Subject} & \multicolumn{1}{c}{\textbf{Textbook}} \\ 
\midrule

\multirow{2}{*}{Astronomy} & 1.\hspace{0.5em} \emph{Foundations of Astrophysics } \cite{ryden2020foundations} \\
& 2.\hspace{0.5em} \emph{Stellar Structure And Evolution } \cite{pols2011stellar} \\
\midrule

\multirow{5}{*}{Biology} & 1.\hspace{0.5em} \emph{Biology, 2nd Edition } \cite{clark2018biology} \\
& 2.\hspace{0.5em} \emph{Introduction to Agricultural Engineering Technology: A Problem Solving Approach, 4th Edition } \cite{field2018introduction} \\
& 3.\hspace{0.5em} \emph{Introduction to Environmental Engineering, 5th Edition } \cite{davis2012introduction} \\
& 4.\hspace{0.5em} \emph{The Economy of Nature, 7th Edition } \cite{ricklefs2013economy} \\
& 5.\hspace{0.5em} \emph{The Molecular Biology of the Cell, 6th Edition } \cite{alberts2014molecular} \\
\midrule

\multirow{5}{*}{Chemistry} & 1.\hspace{0.5em} \emph{Atkins' Physical Chemistry, 12th Edition } \cite{atkins2023atkins} \\
& 2.\hspace{0.5em} \emph{Chemistry, 2nd Edition } \cite{flowers2019chemistry} \\
& 3.\hspace{0.5em} \emph{Chemistry: The Central Science, 15th Edition } \cite{brown2023chemistry} \\
& 4.\hspace{0.5em} \emph{Organic Chemistry As A Second Language } \cite{klein2024organic} \\
& 5.\hspace{0.5em} \emph{Organic Chemistry, 2nd Edition } \cite{clayden2012organic} \\
\midrule

\multirow{2}{*}{Electromagnetism} & 1.\hspace{0.5em} \emph{Introduction to Electrodynamics, 4th Edition } \cite{griffiths2023introduction} \\
& 2.\hspace{0.5em} \emph{University Physics Volume 2 (Electromagnetism)} \cite{ling2016university2} \\
\midrule

\multirow{3}{*}{Geography} & 1.\hspace{0.5em} \emph{Fundamentals of Geophysics, 2nd Edition } \cite{lowrie2020fundamentals} \\
& 2.\hspace{0.5em} \emph{Human Geography, 12th Edition } \cite{fouberg2020human} \\
& 3.\hspace{0.5em} \emph{Physical Geography: A Landscape Appreciation, 10th Edition } \cite{hess2021physical} \\
\midrule

\multirow{1}{*}{Mechanics} & 1.\hspace{0.5em} \emph{University Physics Volume 1} \cite{ling2016university1} \newline \\
\midrule

\multirow{1}{*}{Modern Physics} & 1.\hspace{0.5em} \emph{University Physics Volume 3} \cite{ling2016university3} \newline \\
\midrule

\multirow{3}{*}{Neurobiology} & 1.\hspace{0.5em} \emph{Neuroscience, 6th Edition } \cite{purves2018neuroscience} \\
& 2.\hspace{0.5em} \emph{Principles of Neural Science, 6th Edition } \cite{kandel2021principles} \\
& 3.\hspace{0.5em} \emph{Principles of Neurobiology } \cite{luo2020principles} \\
\midrule
\multirow{2}{*}{Thermodynamics} & 1.\hspace{0.5em} \emph{An Introduction to Thermal Physics } \cite{schroeder2020introduction} \\
& 2.\hspace{0.5em} \emph{University Physics Volume 2 (Thermodynamics)} \cite{ling2016university2} \\

\bottomrule
\end{tabular}
\caption{
List of textbooks and corresponding example numbers for the \textbf{Science} discipline.
}
\label{tab:science}
\end{table*}

\begin{table*}[h]
\centering
\small
\renewcommand{\arraystretch}{1.05}
\begin{tabular}{p{2.5cm}p{10cm}}
\toprule
\textbf{Subject} & \multicolumn{1}{c}{\textbf{Textbook}} \\ 
\midrule

\multirow{4}{*}{\makecell[l]{Biomedical \\ Engineering \\ }} & 1.\hspace{0.5em} \emph{Biomaterials Science: An Introduction to Materials in Medicine, 4th Edition } \cite{wagner2020biomaterials} \\
& 2.\hspace{0.5em} \emph{Biomaterials and Biopolymers } \cite{domb2023biomaterials} \\
& 3.\hspace{0.5em} \emph{Fundamentals and Advances in Medical Biotechnology } \cite{anwar2022fundamentals} \\
& 4.\hspace{0.5em} \emph{Introduction to Biomedical Engineering, 4th Edition } \cite{enderle2017biomedical} \\
\midrule

\multirow{3}{*}{Civil Engineering} & 1.\hspace{0.5em} \emph{Engineering Geology and Construction } \cite{bell2004engineering} \\
& 2.\hspace{0.5em} \emph{Principles of Geotechnical Engineering, 9th Edition } \cite{das2017principles} \\
& 3.\hspace{0.5em} \emph{Structure for Architects: A Case Study in Steel, Wood, and Reinforced Concrete Design } \cite{bedi2019structure} \\
\midrule

\multirow{7}{*}{Computer Science} & 1.\hspace{0.5em} \emph{Algorithms, 4th Edition } \cite{sedgewick2011algorithms} \\
& 2.\hspace{0.5em} \emph{Computer Organization and Design: The Hardware/Software Interface, 6th Edition } \cite{patterson2022computer} \\
& 3.\hspace{0.5em} \emph{Computer Systems: A Programmer's Perspective, 3rd Edition } \cite{bryant2011computer} \\
& 4.\hspace{0.5em} \emph{Deep Learning } \cite{goodfellow2016deep} \\
& 5.\hspace{0.5em} \emph{Digital Image Processing, 4th Edition } \cite{rafael2018digital} \\
& 6.\hspace{0.5em} \emph{Introduction to Algorithms, 4th Edition } \cite{cormen2022introduction} \\
& 7.\hspace{0.5em} \emph{Operating System Concepts, 10th Edition } \cite{silberschatz2018operating} \\
\midrule

\multirow{2}{*}{\makecell[l]{Electrical \\Engineering \\ }} & 1.\hspace{0.5em} \emph{Electrical Engineering: Principles and Applications, 7th Edition } \cite{hambley2018electrical} \newline \newline \\
\midrule

\multirow{3}{*}{\makecell[l]{Electronics \\and Communication \\}} & 1.\hspace{0.5em} \emph{CMOS Analog Circuit Design, 3rd Edition } \cite{allen2011cmos} \\
& 2.\hspace{0.5em} \emph{Introduction to Communication Systems } \cite{madhow2014introduction} \\
& 3.\hspace{0.5em} \emph{The Art of Electronics, 3rd Edition } \cite{horowitz2015art} \\
\midrule

\multirow{4}{*}{Materials Science} & 1.\hspace{0.5em} \emph{Composite Materials: Science and Engineering, 3rd Edition } \cite{chawla2012composite} \\
& 2.\hspace{0.5em} \emph{Convection in Porous Media, 5th Edition } \cite{nield2017convection} \\
& 3.\hspace{0.5em} \emph{Fiber-Reinforced Composites Materials, Manufacturing, and Design, 3rd Edition } \cite{mallick2007fiber} \\
& 4.\hspace{0.5em} \emph{Materials Science and Engineering: An Introduction, 10th Edition } \cite{callister2020materials} \\
\midrule

\multirow{6}{*}{\makecell[l]{Mechanical \\Engineering \\ }} & 1.\hspace{0.5em} \emph{Industrial Automation: An Engineering Approach}  \\
& 2.\hspace{0.5em} \emph{Industrial Robotics Control: Mathematical Models, Software Architecture, and Electronics Design } \cite{frigeni2022industrial} \\
& 3.\hspace{0.5em} \emph{Intelligent Manufacturing System and Intelligent Workshop } \cite{wangintelligent} \\
& 4.\hspace{0.5em} \emph{Machine Tool Practices, 11th Edition } \cite{kibbe2019machine} \\
& 5.\hspace{0.5em} \emph{Marks' Standard Handbook for Mechanical Engineers, 12th Edition } \cite{avallone2018marks} \\
& 6.\hspace{0.5em} \emph{Modern Control Engineering, 5th Edition } \cite{ogata2010modern} \\

\bottomrule
\end{tabular}
\caption{
List of textbooks and corresponding example numbers for the \textbf{Engineering} discipline.
}
\label{tab:tech__engineering}
\end{table*}

\begin{table*}[h]
\centering
\small
\renewcommand{\arraystretch}{1.05}
\begin{tabular}{p{2.5cm}p{10cm}}
\toprule
\textbf{Subject} & \multicolumn{1}{c}{\textbf{Textbook}} \\ 
\midrule

\multirow{3}{*}{Basic Medicine} & 1.\hspace{0.5em} \emph{Kuby Immunology, 8th Edition } \cite{owen2018kuby} \\
& 2.\hspace{0.5em} \emph{Robbins and Cotran Pathologic Basis of Disease, 10th Edition } \cite{kumar2020robbins} \\
& 3.\hspace{0.5em} \emph{Tissue Barriers in Disease, Injury and Regeneration } \cite{gorbunov2022tissue} \\
\midrule

\multirow{2}{*}{Clinical Medicine} & 1.\hspace{0.5em} \emph{Cecil Essentials of Medicine, 10th Edition } \cite{wing2021cecil} \\
& 2.\hspace{0.5em} \emph{Kumar and Clark's Clinical Medicine, 10th Edition } \cite{feather2020kumar} \\
\midrule

\multirow{1}{*}{Dentistry} & 1.\hspace{0.5em} \emph{Pharmacology and Therapeutics for Dentistry, 7th Edition } \cite{yagiela2010pharmacology} \newline \\
\midrule

\multirow{1}{*}{Pharmacy} & 1.\hspace{0.5em} \emph{The Pharmacological Basis of Therapeutics, 13th Edition } \cite{brunton2017goodman} \newline \\
\midrule

\multirow{3}{*}{\makecell[l]{Preventive \\Medicine \\ }} & 1.\hspace{0.5em} \emph{Public Health and Preventive Medicine, 15th Edition } \cite{maxcy2008maxcy} \\
& \\
& \\

\bottomrule
\end{tabular}
\caption{
List of textbooks and corresponding example numbers for the \textbf{Healthcare} discipline.
}
\label{tab:health__medicine}
\end{table*}

\begin{table*}[h]
\centering
\small
\renewcommand{\arraystretch}{1.05}
\begin{tabular}{p{2.5cm}p{10cm}}
\toprule
\textbf{Subject} & \multicolumn{1}{c}{\textbf{Textbook}} \\ 
\midrule

\multirow{3}{*}{Art} & 1.\hspace{0.5em} \emph{Art Through the Ages: A Global History Volume I, 16th Edition } \cite{kleiner2020art} \\
& 2.\hspace{0.5em} \emph{Introduction to Film Studies, 5th Edition } \cite{nelmes2012introduction} \\
& 3.\hspace{0.5em} \emph{The Filmmaker's Handbook: A Comprehensive Guide for the Digital Age, 5th Edition } \cite{ascher2012filmmaker} \\
\midrule

\multirow{5}{*}{Economics} & 1.\hspace{0.5em} \emph{Intermediate Microeconomics: A Modern Approach, 8th Edition } \cite{varian2010intermediate} \\
& 2.\hspace{0.5em} \emph{Land Resource Economics and Sustainable Development: Economic Policies and the Common Good } \cite{van2011land} \\
& 3.\hspace{0.5em} \emph{Macroeconomics, 9th Edition } \cite{blanchard2024macroeconomics} \\
& 4.\hspace{0.5em} \emph{Principles of Economics, 3rd Edition } \cite{greenlaw2023principles} \\
& 5.\hspace{0.5em} \emph{Principles of Microeconomics, 9th Edition } \cite{mankiw2020principles} \\
\midrule

\multirow{2}{*}{History} & 1.\hspace{0.5em} \emph{Archaeology: Theories Methods and Practice, 7th Edition} \cite{renfrew2016archaeology} \\
& 2.\hspace{0.5em} \emph{World History Volume 1: to 1500 } \cite{kordas2022world} \\
\midrule

\multirow{3}{*}{Law} & 1.\hspace{0.5em} \emph{Arbitration Awards: A Practical Approach } \cite{turner2008arbitration} \\
& 2.\hspace{0.5em} \emph{Contract Law} \cite{turner2013contract} \\
& 3.\hspace{0.5em} \emph{The CISG: A new textbook for students and practitioners} \cite{huber2009cisg} \\
\midrule

\multirow{2}{*}{Literature} & 1.\hspace{0.5em} \emph{An Introduction to Language, 11th Edition} \cite{fromkin2017introduction} \\
& 2.\hspace{0.5em} \emph{The Cambridge Introduction to the Novel} \cite{mackay2010cambridge} \\
\midrule

\multirow{2}{*}{Management} & 1.\hspace{0.5em} \emph{Principles of Management } \cite{bright2019principles} \\
& \\
\bottomrule
\end{tabular}
\caption{
List of textbooks and corresponding example numbers for the \textbf{Humanities and Social Science} discipline.
}
\label{tab:humanities__social_sciences}
\end{table*}

%% file: appendix/A-1-annotation_protocol.tex
\subsection{Annotation Guideline and Interface}\label{app:anno-guidline}\label{app:data_annotation_protocol}
With the goal of ensure the high quality of data, \ours adheres to the following four benchmark construction desiderata, we develop the following annotation interface based on Turkle~\cite{turkle}, an
open-source clone of Amazon's Mechanical Turk: 

\begin{figure}[h]
    \centering
    \includegraphics[width=0.95\textwidth]{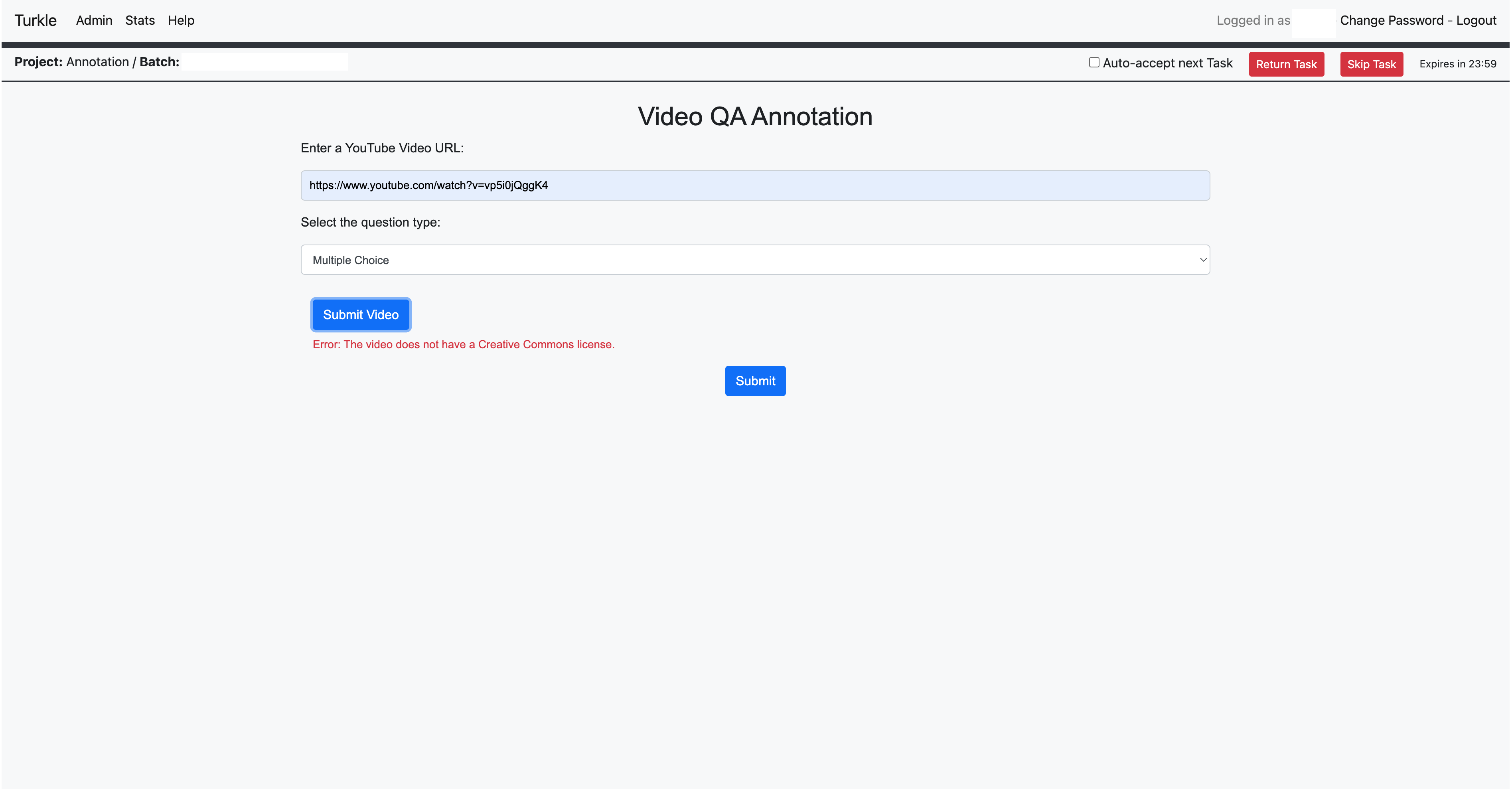}
    \captionof{figure}{
    \textbf{Annotation Interface - Step 1: Video Collection}.
    In this step, annotators are required to input the YouTube video URL and select the desired question type. The backend system of the interface will automatically verify whether the provided YouTube video is under a Creative Commons license using the YouTube Data API v3. If the video does not meet this requirement, as shown in the figure, a warning message will be displayed, and the submission will be blocked. Once a valid example is submitted, the annotation interface will proceed to Step 2, which is illustrated in the following two figures.
    }
\end{figure}

\begin{figure}[h]
    \centering
    \includegraphics[width=0.95\textwidth]{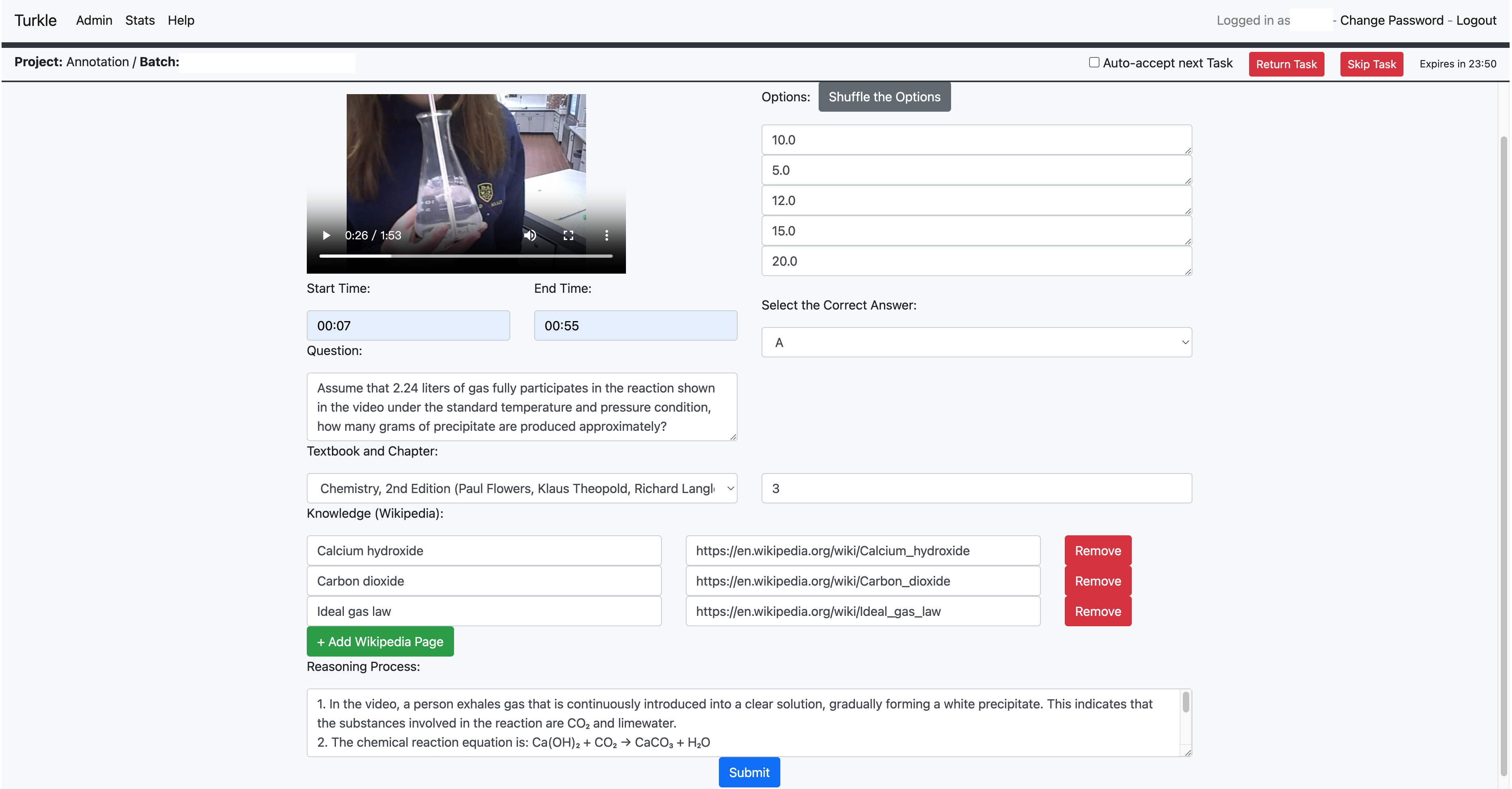}
    \captionof{figure}{
Annotation Interface - Step 2: Multiple-choice Question Annotation.
}

\label{fig:interface-multi}
\end{figure}

\begin{figure}[h]
\centering
\includegraphics[width=0.95\textwidth]{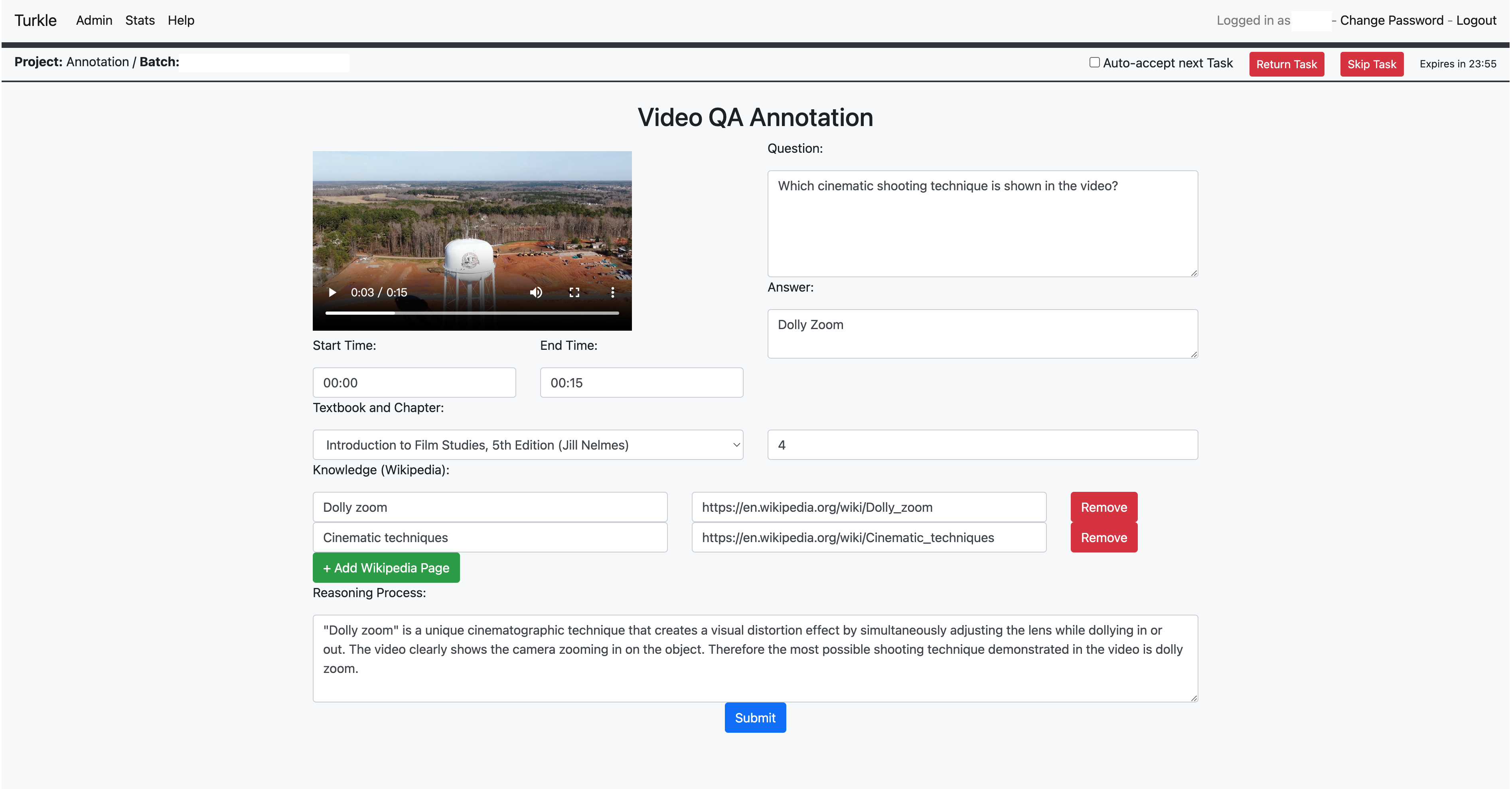}
\captionof{figure}{
Annotation Interface - Step 2: Open-ended Question Annotation.
}
\end{figure}

\clearpage
\onecolumn
\subsection{Validation Guideline and Interface}
To ensure that the final dataset remains high-quality and meets expert-level standards without introducing unnecessary bias, each example in \ours undergoes expert review by one of the authors or top-performing annotators to verify the accuracy of its annotations, following the annotation guideline detailed in \Cref{app:anno-guidline}. The examples of validation interface are presented as follows:

\begin{figure}[h]
\centering
\includegraphics[width=0.95\textwidth]{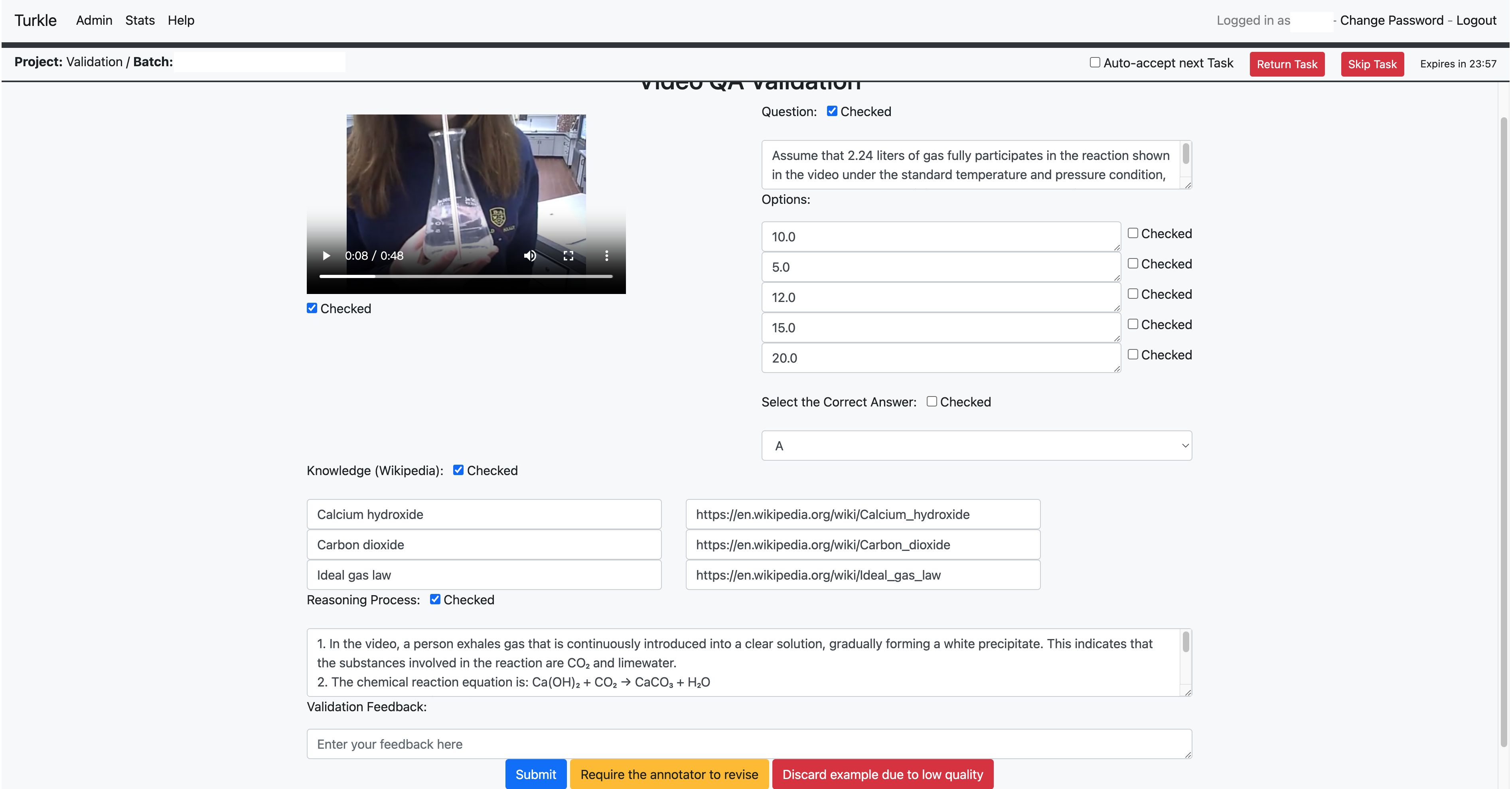}
\captionof{figure}{
\textbf{Validation Interface.}
Human validators are required to thoroughly review each annotation feature to ensure alignment with benchmark construction criteria and annotation guidelines. 
If revisions are not feasible, detailed feedback must be provided to the original annotator, who will then revise and resubmit the annotation for a second review. 
Additionally, validators may discard examples deemed to be of low quality and unlikely to meet the desired criteria through revision.
}
\end{figure}

\subsection{Data Annotation and Validation Payment}\label{app:payment}
The annotation and validation process for \ours spans three months. As outlined in \Cref{sec:data_collection}, annotating examples for \ours can be particularly time-intensive, especially when there is limited availability of videos with Creative Commons licenses in the required subjects. 
To accommodate this and ensure a high-quality dataset, we compensate annotators based on the time they spend rather than the number of examples completed, preventing them from rushing through tasks. 
Annotators are required to record their screens throughout the annotation process, which enables us to verify time reporting accuracy and maintain productivity standards. This also helps us identify any distractions and precisely track the total time spent on each task.
We offer a \emph{base rate} of 6 USD per hour for both annotation and validation work, with an additional 2 USD per completed annotation and 0.40 USD per validated example. 
On average, annotating a single question for \ours takes 20 minutes and 17 seconds, while validation requires 4 minutes and 12 seconds.
This compensation structure ensures that annotators earn wages that are competitive with the average payment for teaching assistants at their respective universities.
To reduce pressure and maintain a comfortable pace, we recommended that annotators limit their work to a maximum of 10 QA example annotations or 50 QA example validations per day.

%% file: appendix/B-experiment_setup.tex
\clearpage
\onecolumn
\section{Experiment Setup}
\subsection{Configuration of Evaluated Models} \label{app:model_info}
\autoref{tab:model_configuration} detail the configuration of each evaluated models.
We use the default settings from the official implementation of each model to process vision input.
Across all experiments, the temperature is set to 1.0, with a maximum output length of 1024 tokens. However, for Gemini-2-Flash-Thinking, the maximum output length is set as 8192 tokens to accommodate its long CoT reasoning mechanism. All inferences are reproducible on a workstation equipped with two NVIDIA A100-80G GPUs.

\input{appendix/tables/B1-model_configuration}

\clearpage
\subsection{Chain-of-Thought and \da Prompts}\label{app:cot_prompt}
The following figures illustrates the CoT reasoning and \da prompts applied in this study for answering multiple-choice and open-ended questions, respectively.

\input{appendix/prompts/2-cot}

\clearpage
\subsection{Prompts for Accuracy Evaluation}\label{app:acc_evaluation_prompt}
\input{appendix/prompts/acc_evaluation}

%% file: appendix/tables/B1-model_configuration.tex
\begin{table*}[h]
\centering
\footnotesize
\resizebox{\textwidth}{!}{%
\begin{tabular}{llllcrcc}
\toprule
\textbf{Organization} & \textbf{Model} & \textbf{Release} & \textbf{Version} & \textbf{\begin{tabular}[c]{@{}c@{}}Support\\Video?\end{tabular}} & \textbf{\begin{tabular}[c]{@{}c@{}}Input\\Frames\end{tabular}} & \textbf{\begin{tabular}[c]{@{}c@{}}\# Inference\\Pipeline\end{tabular}} & \\
\midrule
\multicolumn{8}{c}{\emph{\textbf{Proprietary 
 Models}}} \\
 \midrule
\multirow{3}{*}{OpenAI} & o1$^*$ & 2024-12 & \texttt{o1-2024-12-17} & \xmark & 32 & \multirow{3}{*}{API}\\
& GPT-4o &  2024-8  &  \texttt{gpt-4o-2024-08-06} &\xmark & 32 & \\
& GPT-4o-mini &  2024-7  &  \texttt{gpt-4o-mini-2024-07-18} &\xmark & 32 & \\
\noalign{\vskip 0.5ex}\hdashline\noalign{\vskip 0.5ex}

\multirow{2}{*}{Google} & Gemini 2.0 Flash Thinking & 2024-12 & \texttt{gemini-2.0-flash-thinking-exp-1219} & \xmark & 32 & \multirow{4}{*}{API}\\
& Gemini 2.0 Flash & 2024-12 & \texttt{gemini-2.0-flash-exp} & \xmark & 32 & \\
& Gemini 1.5 Pro & 2024-9 & \texttt{gemini-1.5-pro} &\cmark & 32 & \\
& Gemini 1.5 Flash & 2024-9 & \texttt{gemini-1.5-flash} &\cmark & 32 & &  \\
\noalign{\vskip 0.5ex}\hdashline\noalign{\vskip 0.5ex}

\multirow{1}{*}{Anthropic} & Claude-3.5-Sonnet & 2024-10 & \texttt{claude-3-5-sonnet-20241022} &\xmark & 32 & API\\
\noalign{\vskip 0.5ex}\hdashline\noalign{\vskip 0.5ex}

\multirow{1}{*}{xAI} & Grok-2-Vision & 2024-12 & \texttt{grok-2-vision-1212} & \xmark & 32 & API\\
\noalign{\vskip 0.5ex}\hdashline\noalign{\vskip 0.5ex}

\multirow{1}{*}{Zhipu AI} & GLM-4V-Plus & 2025-1 & \texttt{glm-4v-plus-0111} & \cmark  & 4 & API\\

\midrule
\multicolumn{8}{c}{\emph{\textbf{Open-source Multimodal Foundation Models}}} \\
 \midrule

\multirow{1}{*}{Mistral AI} & Pixtral-12B & 2024-9 & \texttt{Pixtral-12B-2409} &\xmark & 8 & vLLM\\
\noalign{\vskip 0.5ex}\hdashline\noalign{\vskip 0.5ex}

\multirow{1}{*}{Microsoft} & Phi-3.5-Vision & 2024-7 & \texttt{Phi-3.5-vision-instruct} &\xmark & 16 & vLLM\\
\noalign{\vskip 0.5ex}\hdashline\noalign{\vskip 0.5ex}

\multirow{3}{*}{Shanghai AI Lab} 
& InternVL2.5-38B & 2024-11 & \texttt{InternVL2.5-38B} & \xmark & 4 & \multirow{3}{*}{vLLM}\\
& InternVL2.5-8B & 2024-11 & \texttt{InternVL2.5-8B} & \xmark & 4 & \\
& InternVL2-8B & 2024-6 & \texttt{InternVL2-8B} &\xmark & 4 & \\
\noalign{\vskip 0.5ex}\hdashline\noalign{\vskip 0.5ex}

\multirow{3}{*}{Alibaba} & Qwen2-VL-2B & 2024-8 & \texttt{Qwen2-VL-2B-Instruct} &\cmark & 1fps & \multirow{3}{*}{vLLM}\\
& Qwen2-VL-7B & 2024-8 & \texttt{Qwen2-VL-7B-Instruct} &\cmark & 1fps & \\
& Qwen2-VL-72B & 2024-9 & \texttt{Qwen2-VL-72B-Instruct} &\cmark & 1fps & \\
\noalign{\vskip 0.5ex}\hdashline\noalign{\vskip 0.5ex}

\multirow{2}{*}{Meta} & Llama-3.2-11B-Vision & 2024-9 & \texttt{Llama-3.2-11B-Vision-Instruct} &\xmark & 8 & \multirow{2}{*}{vLLM} \\
& Llama-3.2-90B-Vision & 2024-9 & \texttt{Llama-3.2-90B-Vision-Instruct} &\xmark & 8 &  \\
\noalign{\vskip 0.5ex}\hdashline\noalign{\vskip 0.5ex}

\multirow{2}{*}{DAMO} & VideoLLaMA2-7B & 2024-6 & \texttt{VideoLLaMA2-7B} & \cmark & 1fps & HF\\
& VideoLLaMA2.1-7B & 2024-10 & \texttt{
VideoLLaMA2.1-7B-16F} & \cmark & 1fps & HF\\
\noalign{\vskip 0.5ex}\hdashline\noalign{\vskip 0.5ex}

\multirow{3}{*}{DeepSeek} & DeepSeek-VL2 & 2024-12 & \texttt{deepseek-vl2 } & \xmark & 2 & vLLM \\
& DeepSeek-VL2-Small & 2024-12 & \texttt{deepseek-vl2-small} & \xmark & 2 & vLLM\\
& DeepSeek-VL2-Tiny & 2024-12 & \texttt{deepseek-vl2-tiny} & \xmark & 2 & vLLM \\
\noalign{\vskip 0.5ex}\hdashline\noalign{\vskip 0.5ex}

\multirow{1}{*}{Rhymes} & Aria & 2024-11 & \texttt{Aria-Chat} & \xmark & 8 & vLLM \\
\noalign{\vskip 0.5ex}\hdashline\noalign{\vskip 0.5ex}

\multirow{4}{*}{Llava Hugging Face} & LLaVA-OneVision-7B & 2024-9 & \texttt{llava-onevision-qwen2-7b-ov-chat-hf} & \cmark & 1fps & vLLM \\
& LLaVA-NeXT-Video-34B & 2024-6 & \texttt{LLaVA-NeXT-Video-34B-hf} & \xmark & 8 & vLLM \\
& LLaVA-NeXT-Video-7B & 2024-6 & \texttt{LLaVA-NeXT-Video-7B-hf} &\cmark & 16 & vLLM \\
\noalign{\vskip 0.5ex}\hdashline\noalign{\vskip 0.5ex}

\multirow{1}{*}{HuggingFaceM4} & Idefics3-8B & 2024-8 & \texttt{Idefics3-8B-Llama3} & \xmark & 4 & vLLM\\
\noalign{\vskip 0.5ex}\hdashline\noalign{\vskip 0.5ex}

\multirow{1}{*}{OpenGVLab} & InternVideo2-8B & 2024-8 & \texttt{InternVideo2-Chat-8B} & \cmark & 1fps & HF \\
\noalign{\vskip 0.5ex}\hdashline\noalign{\vskip 0.5ex}

\multirow{1}{*}{H2O} & H2OVL Mississippi-2B & 2024-10 & \texttt{h2ovl-mississippi-2b} & \xmark & 4 & vLLM \\

\bottomrule
\end{tabular}
}
\caption{
Details of the multimodal foundation models evaluated in \ours. The ``Source'' column includes URLs for proprietary models and Hugging Face model names for open-source models.
The ``\# Input Frames'' column, for those models only support multi-image input, represents the default number of input frames, chosen from {2, 4, 8, 16, 32}, based on the maximum value that does not exceed the model's context window. ``HF'' means ``Hugging Face''.
}
\label{tab:model_configuration}
\end{table*}

%% file: appendix/prompts/2-cot.tex
\begin{figure}[h]
\centering
\begin{minipage}{\linewidth}
\begin{tcolorbox}[colback=black!7.5!white, colframe=black!30!white, fontupper=\footnotesize, fonttitle=\footnotesize]

Question:\{question\} \\
A: \{option\_a\} \\
B: \{option\_b\} \\
C: \{option\_c\} \\
D: \{option\_d\} \\
E: \{option\_e\} \\

Visual Information: \{processed\_video\} \\

Answer the given multiple-choice question step by step. Begin by explaining your reasoning process clearly. Conclude by stating the final answer using the following format: ``Therefore, the final answer is: \$LETTER'' (without quotes), where \$LETTER is one of the options. Think step by step before answering.
\end{tcolorbox}
\end{minipage}
\caption{CoT reasoning prompt, adopted from MMMU-Pro~\cite{yue2024mmmupro}, for answering multiple-choice question.}
\end{figure}

\begin{figure}[h]
\centering
\begin{minipage}{\linewidth}
\begin{tcolorbox}[colback=black!7.5!white, colframe=black!30!white, fontupper=\footnotesize, fonttitle=\footnotesize]

Question:\{question\} \\

Visual Information: \{processed\_video\} \\

Answer the given question step by step. Begin by explaining your reasoning process clearly. Conclude by stating the final answer using the following format: 'Therefore, the final answer is: ``Answer: \$ANSWER'' (without quotes), where \$ANSWER is the final answer of the question. Think step by step before answering.
\end{tcolorbox}
\end{minipage}
\caption{CoT reasoning prompt for answering open-ended question.}
\end{figure}

\begin{figure}[h]
\centering
\begin{minipage}{\linewidth}
\begin{tcolorbox}[colback=black!7.5!white, colframe=black!30!white, fontupper=\footnotesize, fonttitle=\footnotesize]

Question:\{question\} \\
A: \{option\_a\} \\
B: \{option\_b\} \\
C: \{option\_c\} \\
D: \{option\_d\} \\
E: \{option\_e\} \\

Visual Information: \{processed\_video\} \\

Do not generate any intermediate reasoning process. Answer directly with the option letter from the given choices.
\end{tcolorbox}
\end{minipage}
\caption{\da prompt, adopted from MMMU-Pro~\cite{yue2024mmmupro}, for answering multiple-choice question.}
\end{figure}

\begin{figure}[h]
\centering
\begin{minipage}{\linewidth}
\begin{tcolorbox}[colback=black!7.5!white, colframe=black!30!white, fontupper=\footnotesize, fonttitle=\footnotesize]

Question:\{question\} \\

Visual Information: \{processed\_video\} \\

Do not generate any intermediate reasoning process. Directly output the final answer.
\end{tcolorbox}
\end{minipage}
\caption{\da prompt for answering open-ended question.}
\end{figure}

%% file: appendix/prompts/acc_evaluation.tex
\begin{figure}[h]
\centering
\begin{minipage}{\linewidth}
\begin{tcolorbox}[colback=black!7.5!white, colframe=black!30!white, fontupper=\footnotesize, fonttitle=\footnotesize]
[Instruction]\\
Evaluate whether the model's final answer is correct by comparing it to the ground-truth answer provided for the given question.

You should first extract the final answer from the model's response, and then compare the extracted answer with the ground-truth answer to determine its accuracy.
Output your response in the following structured format:\\
\{\\
\hspace*{1.5em} `extracted\_answer': // str value "A" "B" "C" "D" "E", should be a single character\\
\hspace*{1.5em} `correct': // boolean value, True if the answer is correct, False otherwise\\
\}\\

[User]\\
Question:\{question\} \\
A: \{option\_a\} \\
B: \{option\_b\} \\
C: \{option\_c\} \\
D: \{option\_d\} \\
E: \{option\_e\} \\

Ground Truth Answer: \{ground\_truth\} \\

Model Response to the Question: \{model\_response\}\\
\end{tcolorbox}
\end{minipage}
\caption{Evaluation prompt used for assessing the accuracy of multi-choice QA.}
\end{figure}

\begin{figure}[h]
\centering
\begin{minipage}{\linewidth}
\begin{tcolorbox}[colback=black!7.5!white, colframe=black!30!white, fontupper=\footnotesize, fonttitle=\footnotesize]
[Instruction]\\
Evaluate whether the model's final answer is correct by comparing it to the ground-truth answer provided for the given question.
You should first extract the final answer from the model's response, and then compare the extracted answer with the ground-truth answer to determine its accuracy.
The final answer generated by the model does not need to match the ground-truth answer word-for-word. However, it should only be considered correct if it demonstrates the exact same technique or concept explicitly and unambiguously equivalent to the ground-truth answer.
Output your response in the following structured format:\\
\{\\
\hspace*{1.5em} `extracted\_answer': // str value, the short final answer extracted from the model's response, do not hallucinate one that is not present in the response\\
\hspace*{1.5em} `correct': // boolean value, True if the answer is correct, False otherwise\\
\}\\

[User]\\
Question:\{question\} \\

Ground Truth Answer: \{ground\_truth\} \\

Model Response to the Question: \{model\_response\}\\
\end{tcolorbox}
\end{minipage}
\caption{Evaluation prompt used for assessing the accuracy of open-ended QA.}
\end{figure}

%% file: appendix/E-case_study.tex
\clearpage
\onecolumn
\section{Experiment}\label{app:case_study}
\subsection{Comparison Between CoT Reasoning and Direct Answering}\label{app:cot_compare}
\begin{figure}[h]
 \centering
    \includegraphics[width=0.95\linewidth]{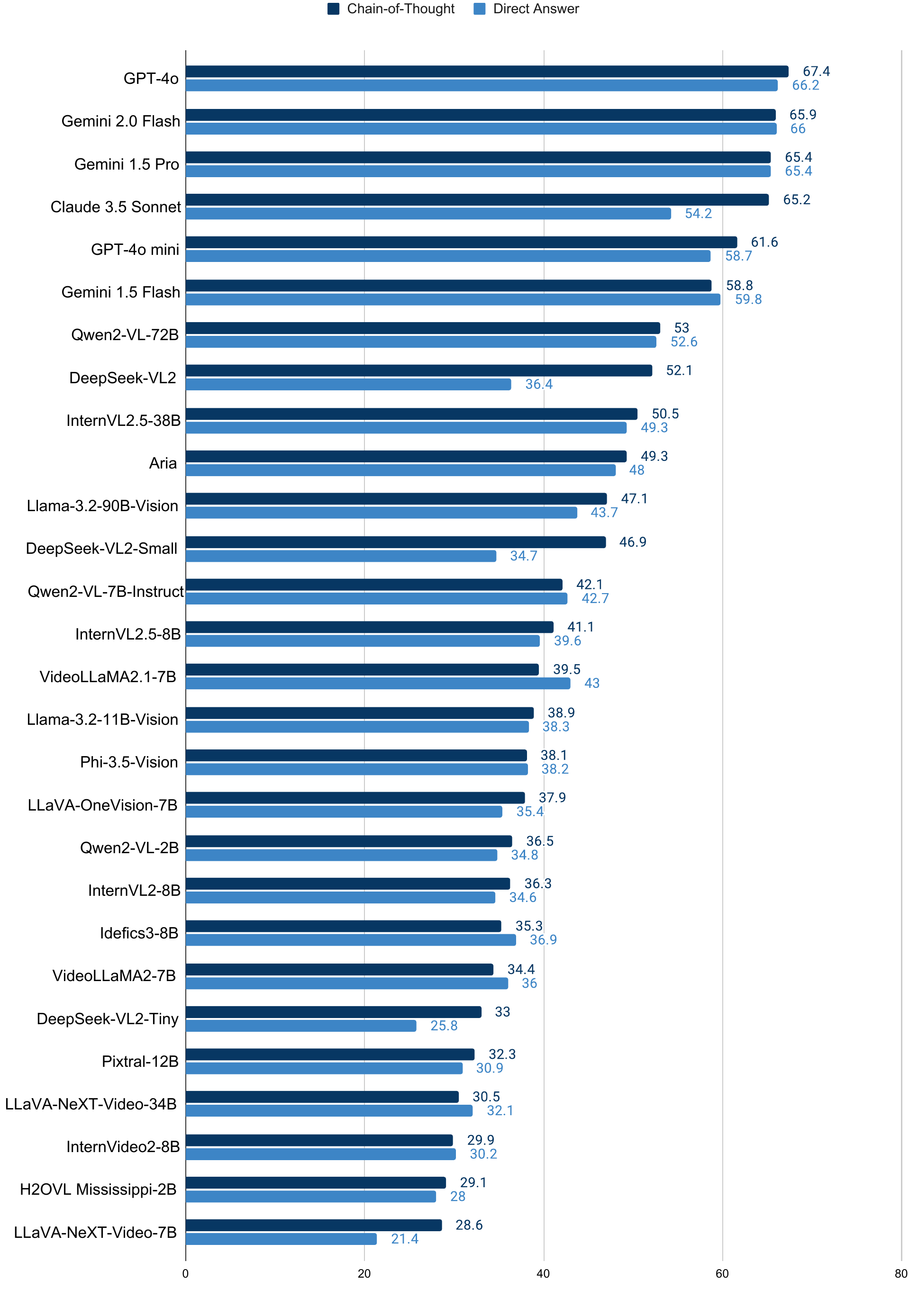}
    \caption{Comparison of model performance between CoT reasoning and direct answering on the validation set.}
    \label{fig:cot_da_compare}
\end{figure}

\clearpage
\subsection{Error Case Analysis: Visual Perception Error}
\begin{figure}[h]
    \centering
    \includegraphics[width=0.9\textwidth]{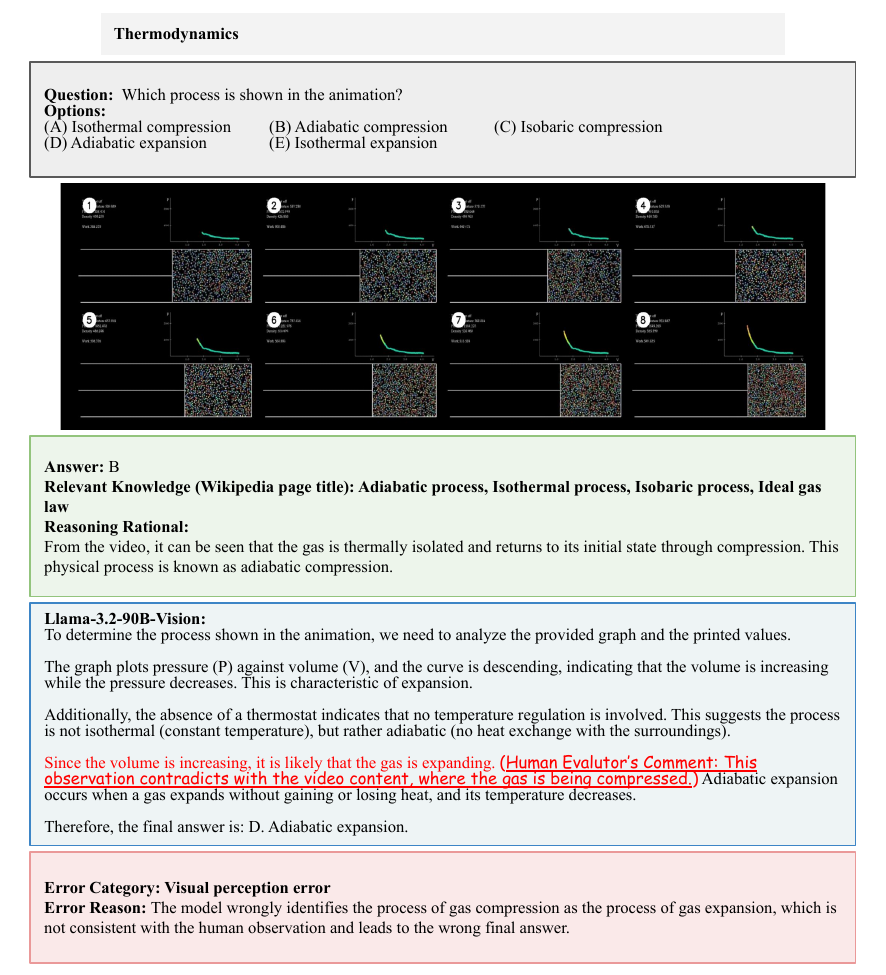}
    \caption{An error case of Thermodynamics.}
    \label{fig:visual_perception_error_1}
\end{figure}
\clearpage

\begin{figure}[h]
    \centering
    \includegraphics[width=0.9\textwidth]{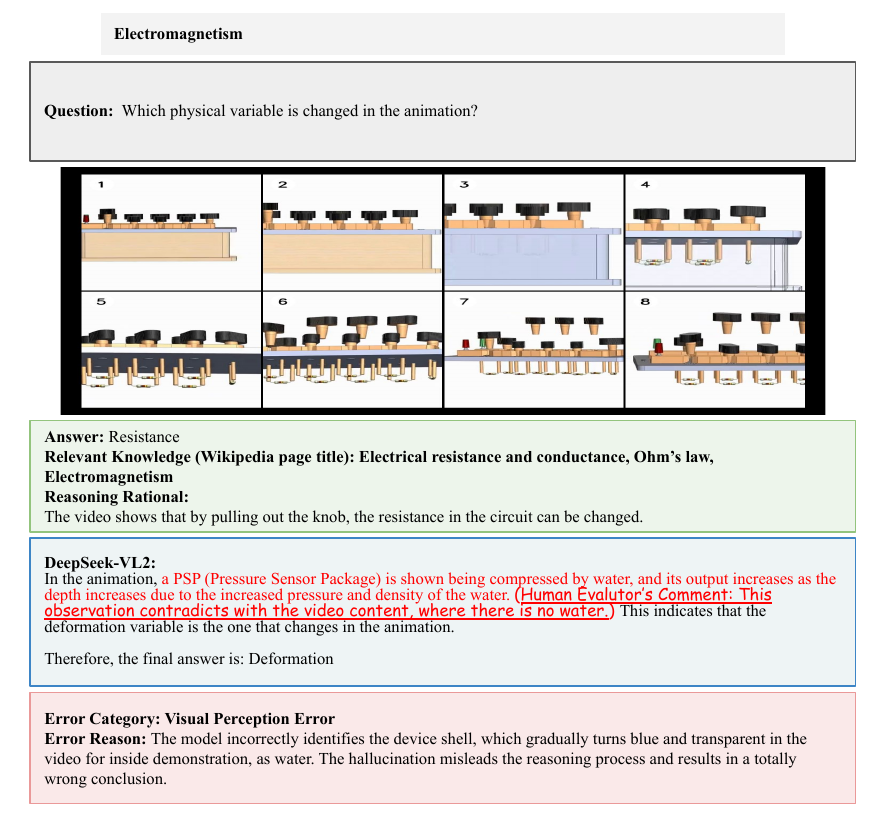}
    \caption{An error case of Electromagnetism.}
    \label{fig:visual_perception_error_2}
\end{figure}
\clearpage

\begin{figure}[h]
    \centering
    \includegraphics[width=0.9\textwidth]{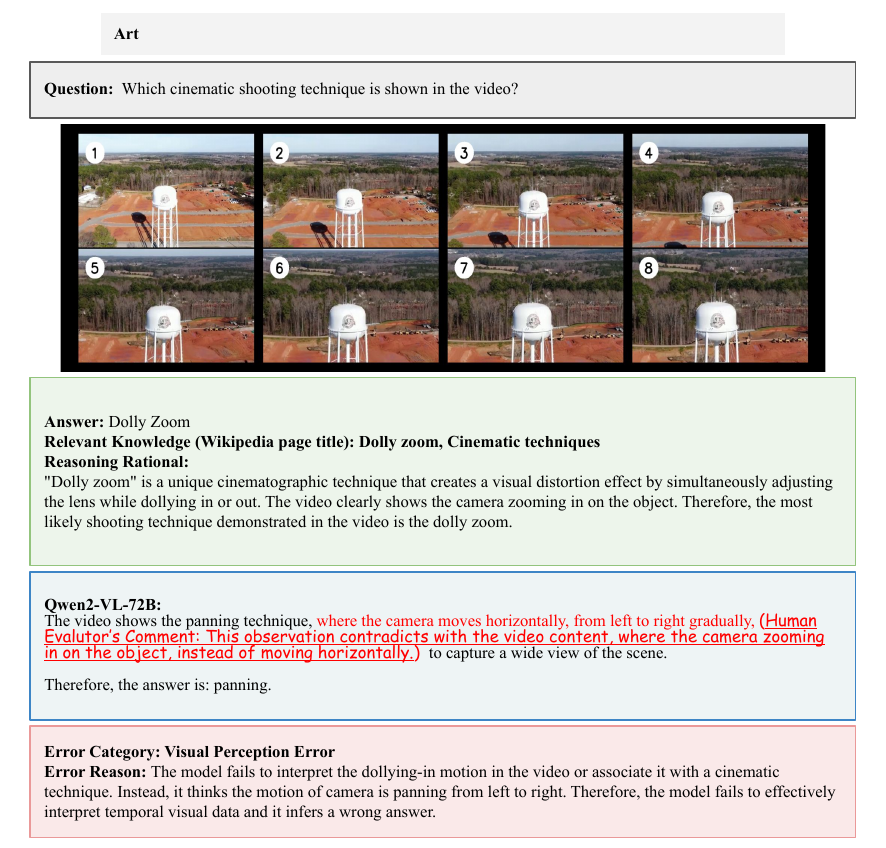}
    \caption{An error case of Art.}
    \label{fig:visual_perception_error_3}
\end{figure}
\clearpage

\subsection{Error Case Analysis: Misuse or Lack Domain Knowledge in Visual Perception}

\begin{figure}[h]
    \centering
    \includegraphics[width=0.9\textwidth]{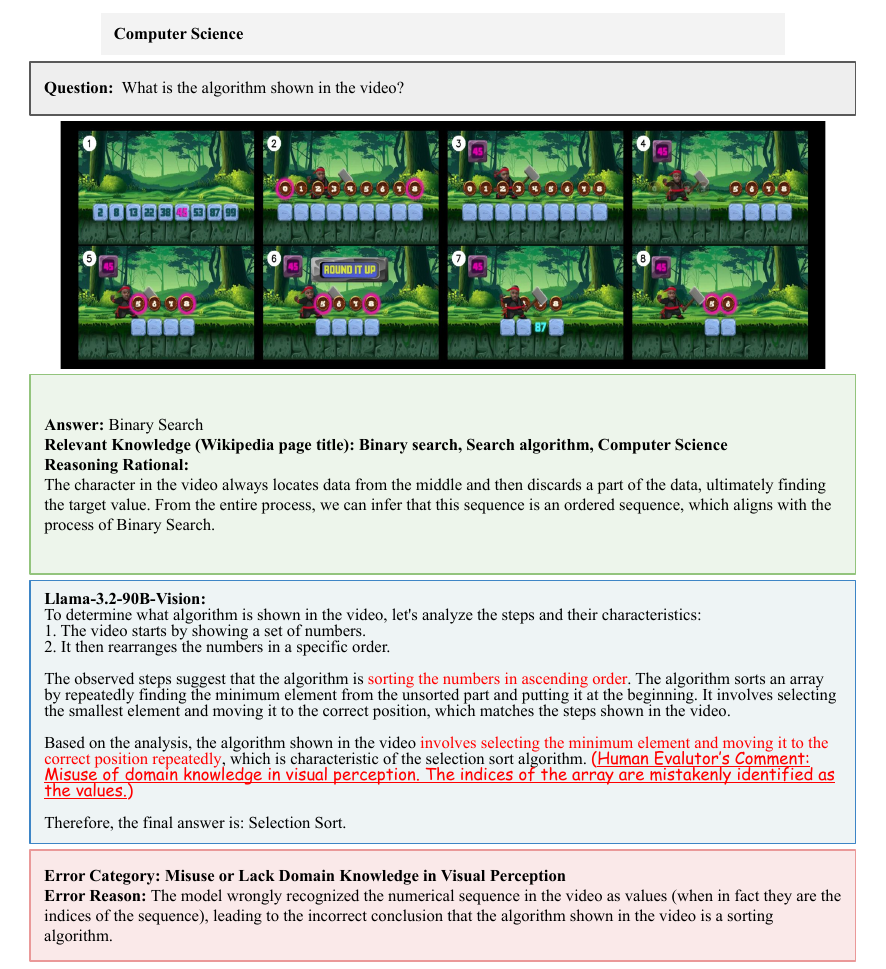}
    \caption{An error case of Computer Science.}
    \label{fig:visual_perception_misuse_1}
\end{figure}
\clearpage

\begin{figure}[h]
    \centering
    \includegraphics[width=0.9\textwidth]{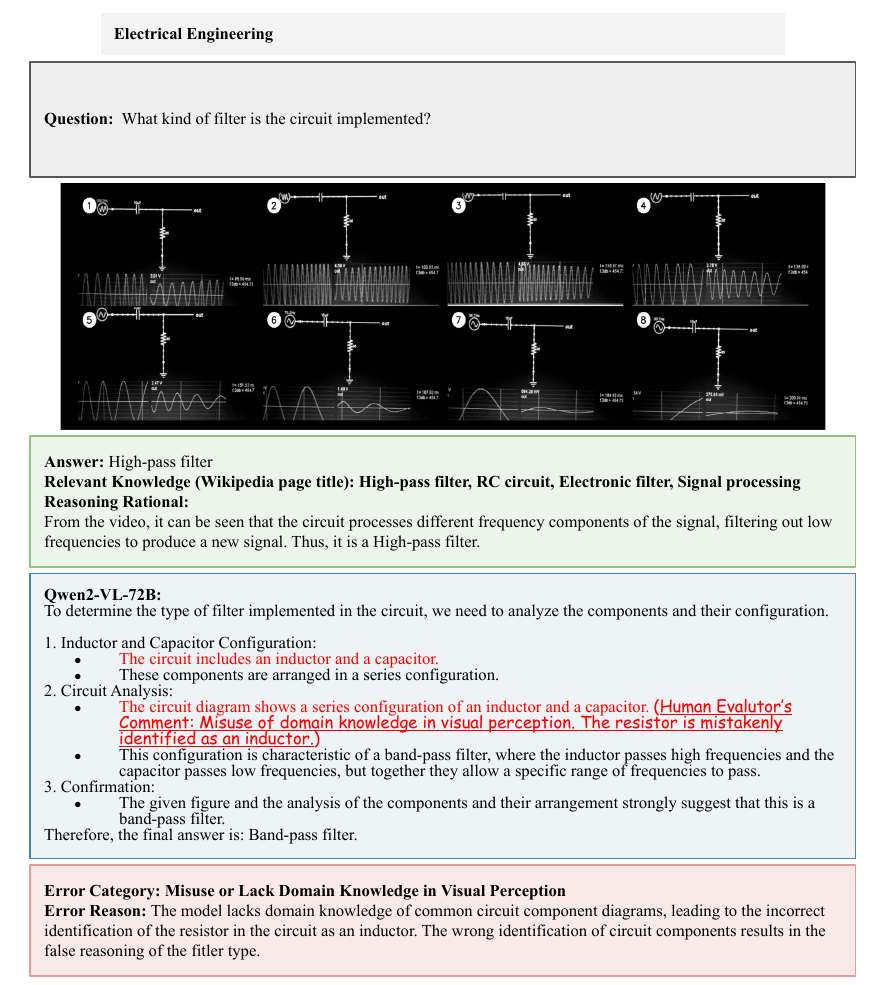}
    \caption{An error case of Electrical Engineering.}
    \label{fig:visual_perception_misuse_2}
\end{figure}
\clearpage

\begin{figure}[h]
    \centering
    \includegraphics[width=0.9\textwidth]{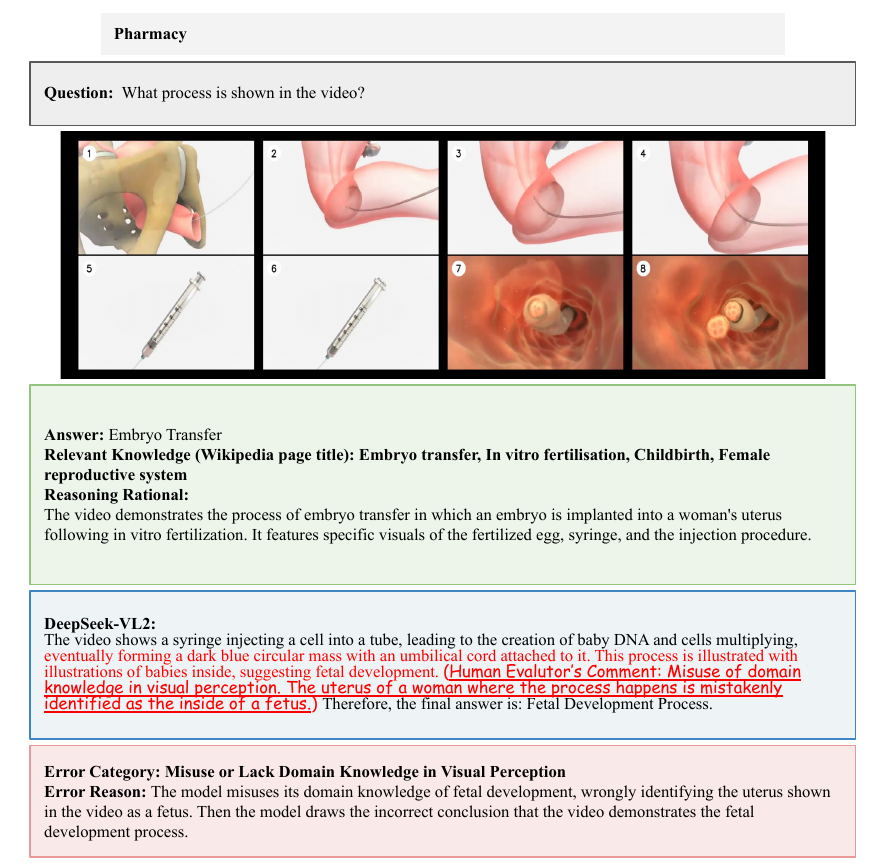}
    \caption{An error case of Pharmacy.}
    \label{fig:visual_perception_misuse_3}
\end{figure}
\clearpage

\subsection{Error Case Analysis: Misuse or Lack Domain Knowledge in Reasoning}

\begin{figure}[h]
    \centering
    \includegraphics[width=0.9\textwidth]{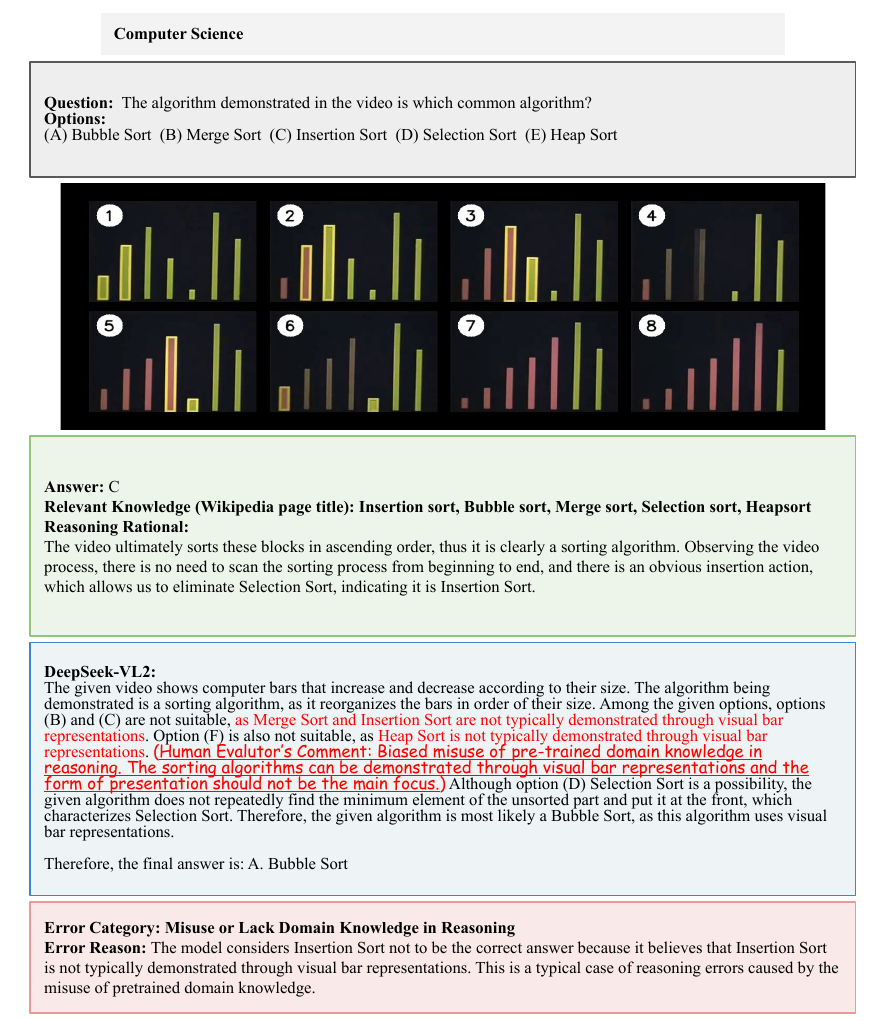}
    \caption{An error case of Computer Science.}
    \label{fig:knowledge_reasoning_1}
\end{figure}
\clearpage

\begin{figure}[h]
    \centering
    \includegraphics[width=0.9\textwidth]{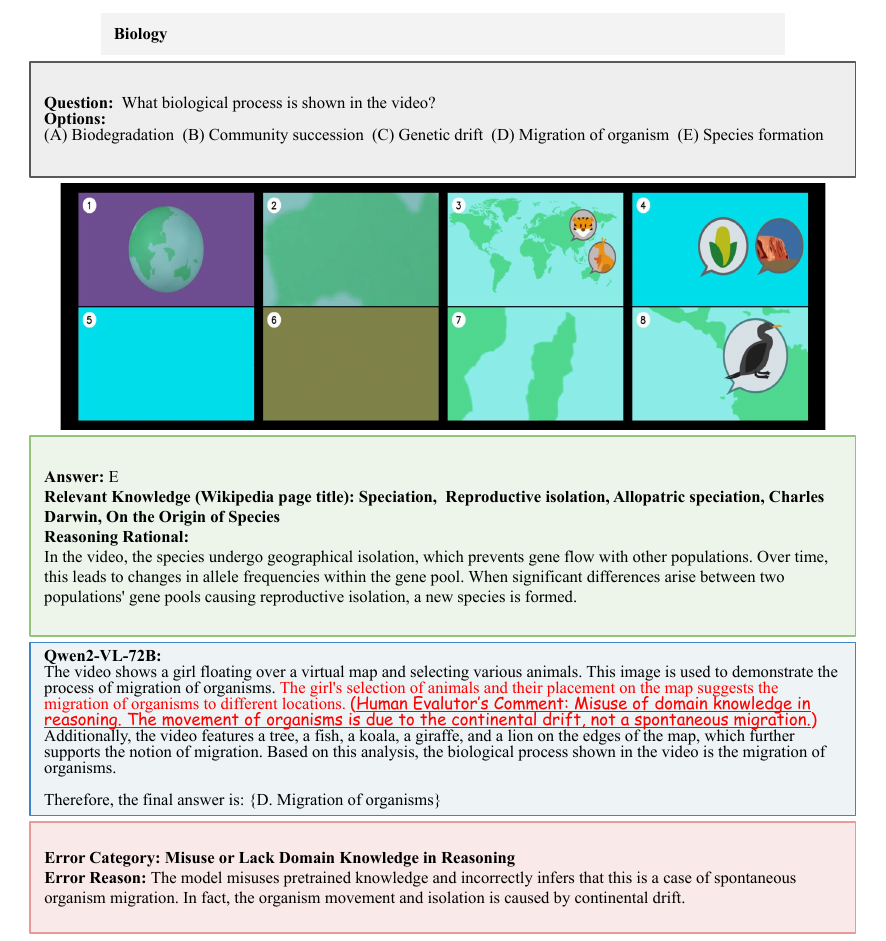}
    \caption{An error case of Biology.}
    \label{fig:knowledge_reasoning_2}
\end{figure}
\clearpage

\begin{figure}[h]
    \centering
    \includegraphics[width=0.9\textwidth]{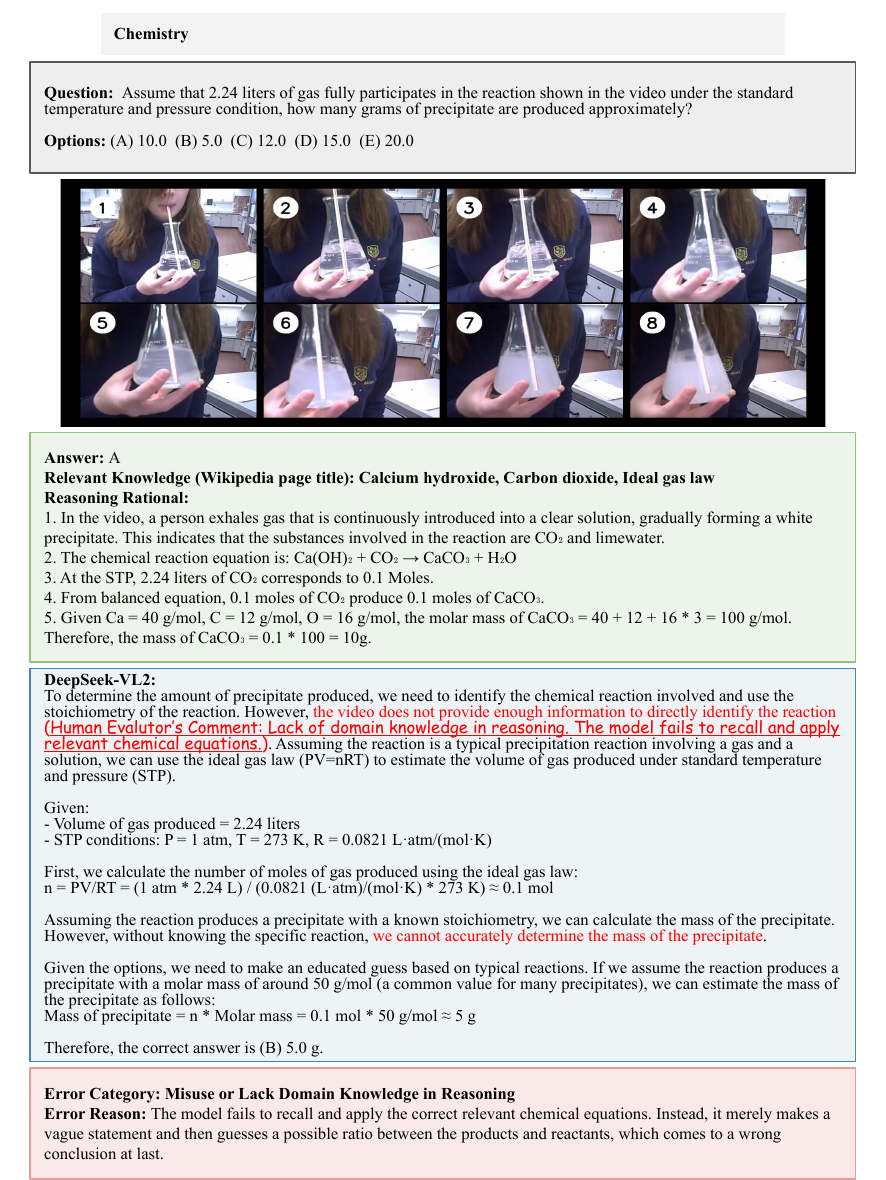}
    \caption{An error case of Chemistry.}
    \label{fig:knowledge_reasoning_3}
\end{figure}
\clearpage

\subsection{Error Case Analysis: Heavy Reliance on Textual Information}

\begin{figure}[h]
    \centering
    \includegraphics[width=0.9\textwidth]{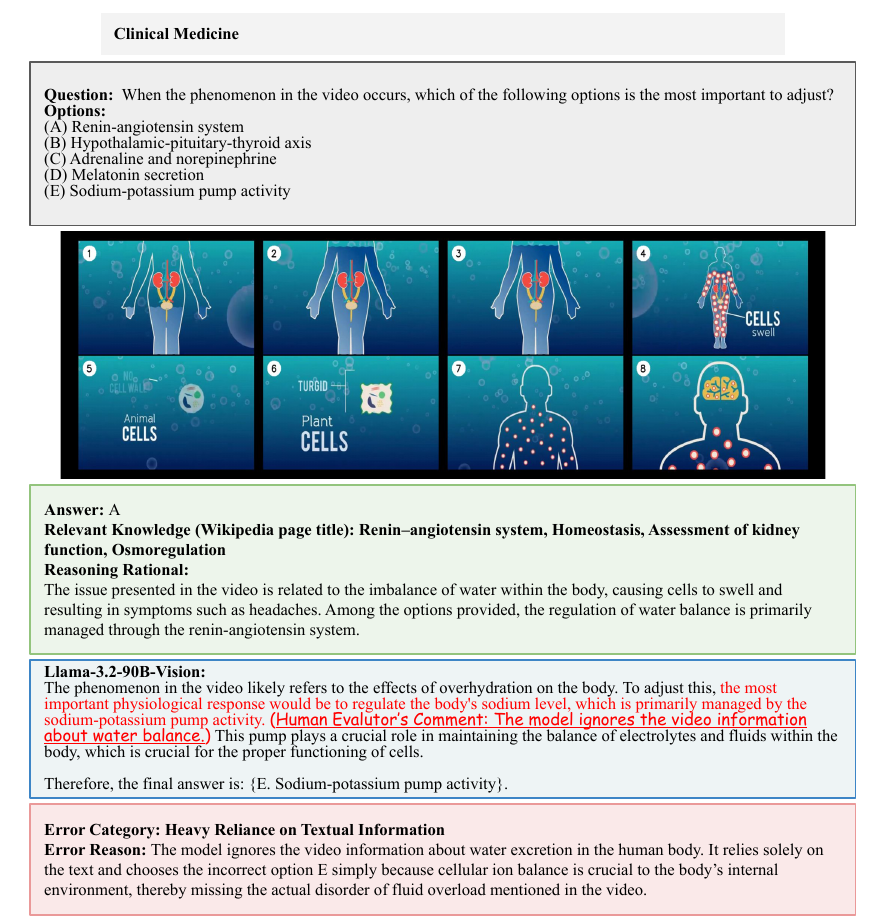}
    \caption{An error case of Clinical Medicine.}
    \label{fig:text_error_1}
\end{figure}
\clearpage

\begin{figure}[h]
    \centering
    \includegraphics[width=0.9\textwidth]{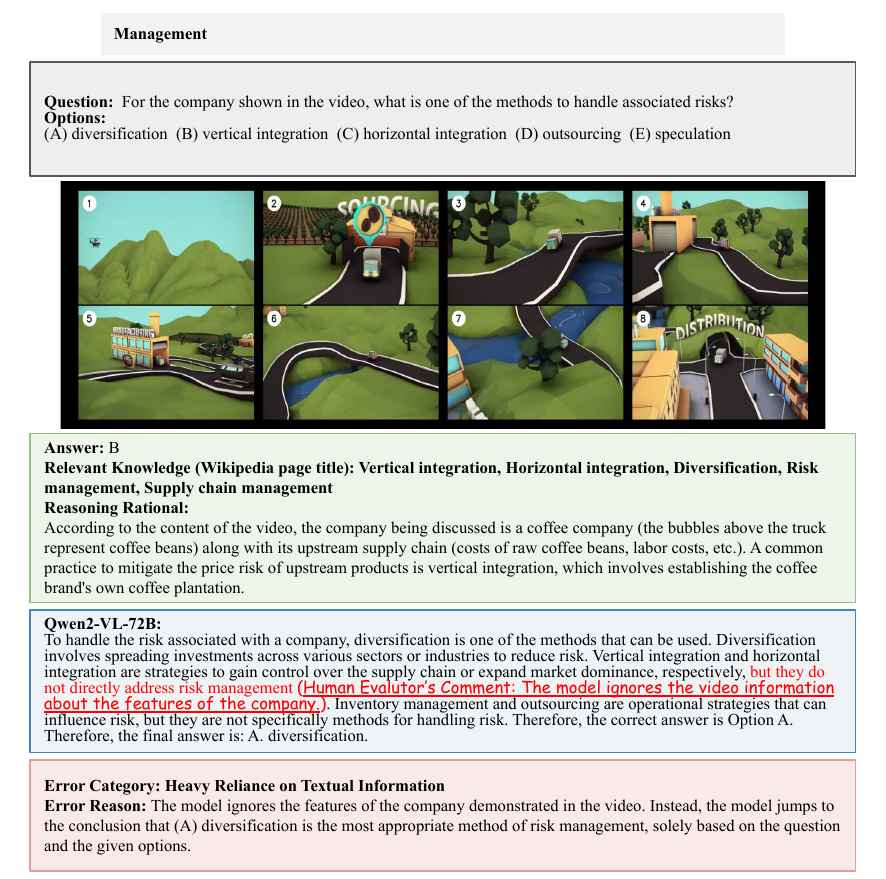}
    \caption{An error case of Management.}
    \label{fig:text_error_2}
\end{figure}
\clearpage

\subsection{Error Case Analysis: Logical Reasoning Error}
\begin{figure}[h]
    \centering
    \includegraphics[width=0.9\textwidth]{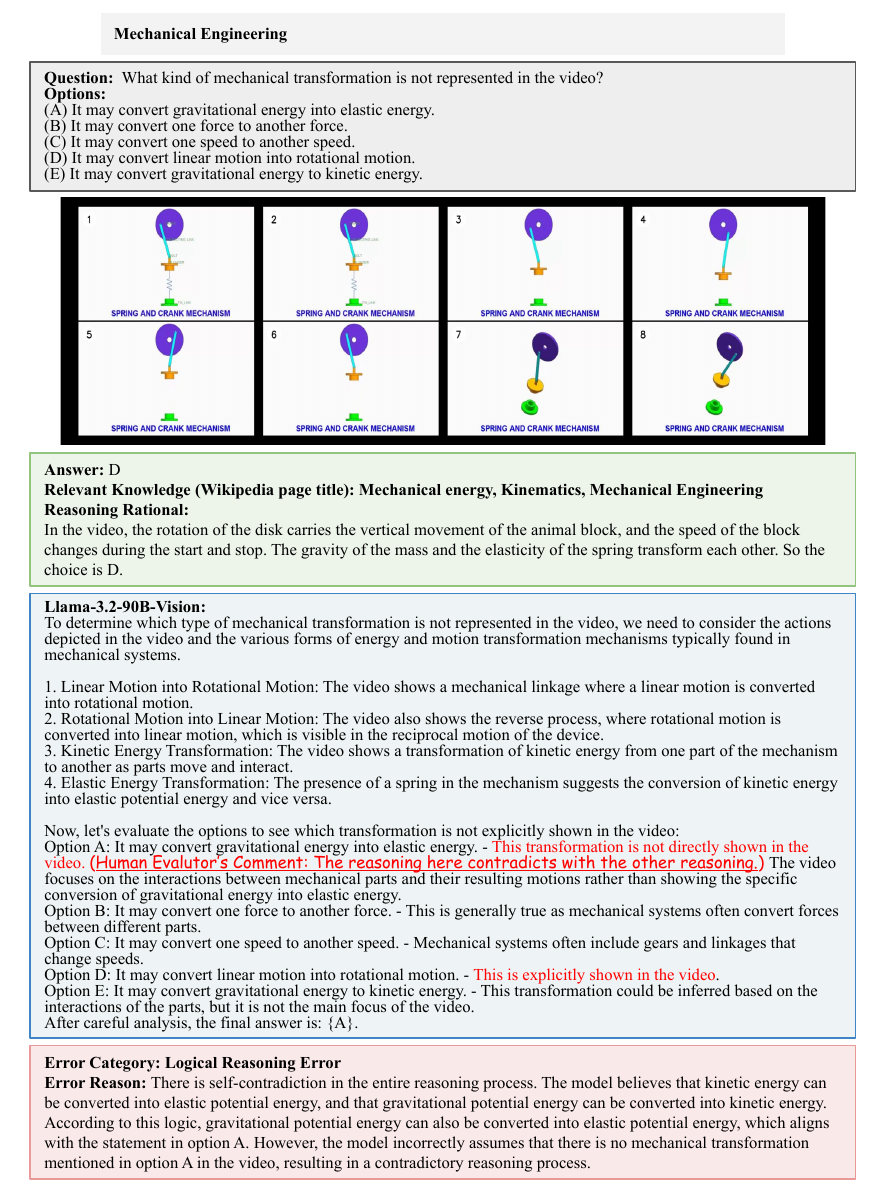}
    \caption{An error case of Mechanical Engineering.}
    \label{fig:logical_error_1}
\end{figure}
\clearpage

\begin{figure}[h]
    \centering
    \includegraphics[width=0.9\textwidth]{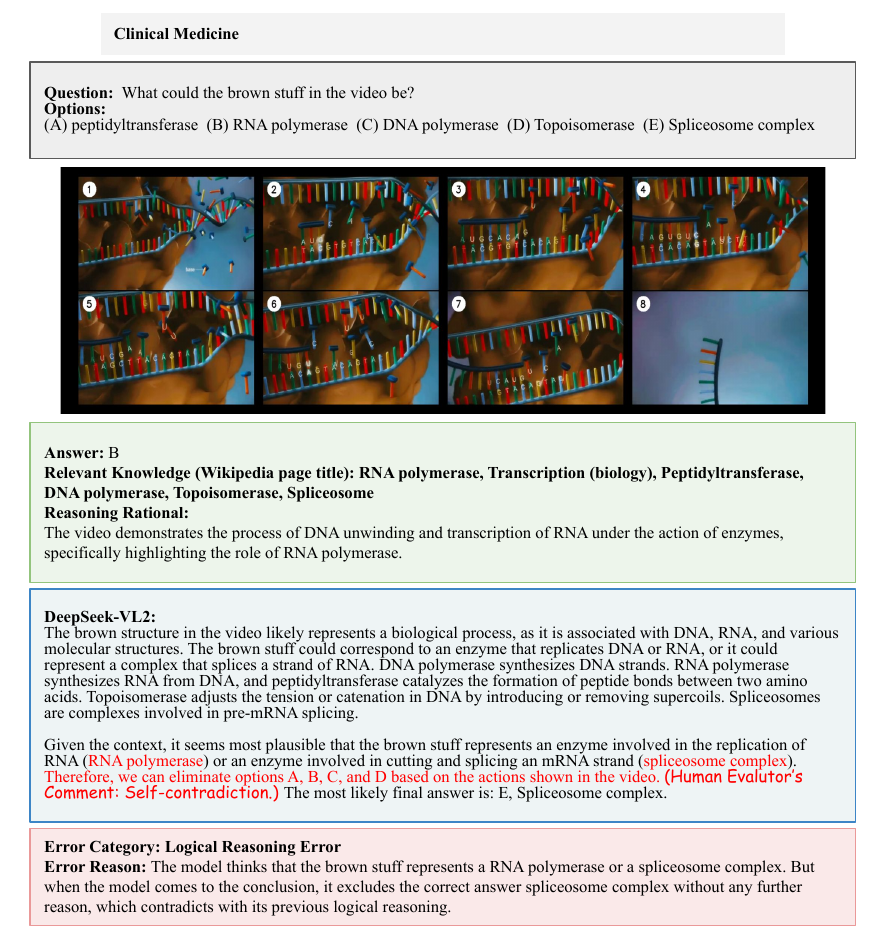}
    \caption{An error case of Clinical Medicine.}
    \label{fig:logical_error_2}
\end{figure}
\clearpage